%% file: _main.tex
\input{_constants}
\arxiv % \review OR \arxiv OR \cameraready

\pdfoutput=1
\documentclass[10pt,twocolumn,letterpaper]{article}
\input{cvpr_header}
\begin{document}
%% TITLE
\title{\paperTitle}
\author{\authorBlock}

\input{figs/teaser}
% \maketitle
%%
%%

\definecolor{myred}{HTML}{bf0511}
\definecolor{myblue}{HTML}{1272b6}

\definecolor{cvprblue}{rgb}{0.21,0.49,0.74}

\input{00_abstract}
\input{01_intro}
\input{02_related}

\input{03_method}
\input{04_experiments}

\input{10_conclusion}

{\small
\bibliographystyle{ieee_fullname}
\bibliography{11_references}
}

\ifarxiv \clearpage \input{12_appendix} \fi

\end{document}

%% file: _constants.tex
\def\paperID{*****}
\def\confName{CVPR}
\def\confYear{2026}

\def\paperTitle{
\SHORTNAME! Expressive Portrait Image Animation for Live Streaming
}

\def\authorBlock{
    Zhiyuan Li$^{1,2,3}$ \qquad
    Chi-Man Pun$^{1,}$\thanks{Equal contribution} \qquad
    Chen Fang$^{2}$ \qquad
    Jue Wang$^{2}$ \qquad
    Xiaodong Cun$^{3,}\footnotemark[1]$  \\ \\
    $^1$~University of Macau \qquad
    $^2$~Dzine.ai \qquad
    $^3$~GVC Lab, Great Bay University
     \\\\ \url{https://github.com/GVCLab/PersonaLive}
}

\newif\ifreview 
\newif\ifarxiv \newcommand{\arxiv}{\arxivtrue}
\newif\ifcamera 
\newif\ifrebuttal 

%% file: cvpr_header.tex
\ifreview \usepackage[review]{cvpr} \fi
\ifarxiv \usepackage[pagenumbers]{cvpr} \fi
\ifrebuttal \usepackage[rebuttal]{cvpr} \fi
\ifcamera \usepackage{cvpr} \fi

\usepackage{graphicx}
\usepackage{amsmath}
\usepackage{amssymb}
\usepackage{booktabs}

\input{_macros}  % Add packages to _macros.tex

\usepackage{xr-hyper}

\makeatletter
\newcommand*{\addFileDependency}[1]{
  \typeout{(#1)}
  \@addtofilelist{#1}
  \IfFileExists{#1}{}{\typeout{No file #1.}}
}

\makeatother

\usepackage[pagebackref,breaklinks,colorlinks]{hyperref}
\usepackage[capitalize]{cleveref}
\crefname{section}{Sec.}{Secs.}
\crefname{table}{Table}{Tables}
\crefname{figure}{Fig.}{Figs.}

\definecolor{alizarin}{rgb}{0.82, 0.1, 0.26}
\hypersetup{
    colorlinks = true,
    linkbordercolor = {white},
    citecolor=cvprblue,
    urlcolor=alizarin
}

%% file: _macros.tex
\usepackage{times}
\usepackage{microtype}
\usepackage{epsfig}
\usepackage[table,xcdraw]{xcolor}
\usepackage{caption}
\usepackage{float}
\usepackage{placeins}
\usepackage{color, colortbl}
\usepackage{stfloats}
\usepackage{enumitem}
\usepackage{tabularx}
\usepackage{xstring}
\usepackage{multirow}
\usepackage{xspace}
\usepackage{url}
\usepackage{subcaption}
\usepackage{xcolor}
\usepackage[hang,flushmargin]{footmisc}

\usepackage{balance}
\usepackage{color}
\usepackage{listings}
\usepackage{array}
\usepackage{bbding}

\ifcamera \usepackage[accsupp]{axessibility} \fi

% ...

\ifarxiv  \fi

\newcommand{\SHORTNAME}{PersonaLive}

\newcommand{\R}[1]{{%
    \textbf{%
        \ifstrequal{#1}{1}{\textcolor{red}{R#1}}{%
        \ifstrequal{#1}{2}{\textcolor{blue}{R#1}}{%
        \ifstrequal{#1}{3}{\textcolor{magenta}{R#1}}{%
        \ifstrequal{#1}{4}{\textcolor{teal}{R#1}}{%
                           \textcolor{cyan}{R#1}%
        }}}}%
    }%
}}

\usepackage{xspace}

\makeatletter
\DeclareRobustCommand\onedot{\futurelet\@let@token\@onedot}
\def\@onedot{\ifx\@let@token.\else.\null\fi\xspace}

\def\eg{\emph{e.g}\onedot} 
\def\ie{\emph{i.e}\onedot}

\makeatother

\usepackage[ruled,vlined]{algorithm2e}
\SetAlgoLined
\SetKwInput{KwIn}{Input}
\SetKwInput{KwOut}{Output}
\SetKwComment{tcp}{/* }{ */}

%% file: figs/teaser.tex
\twocolumn[{
\maketitle
\begin{center}
    \captionsetup{type=figure}
    \vspace{-2em}
\includegraphics[width=1.0\textwidth]{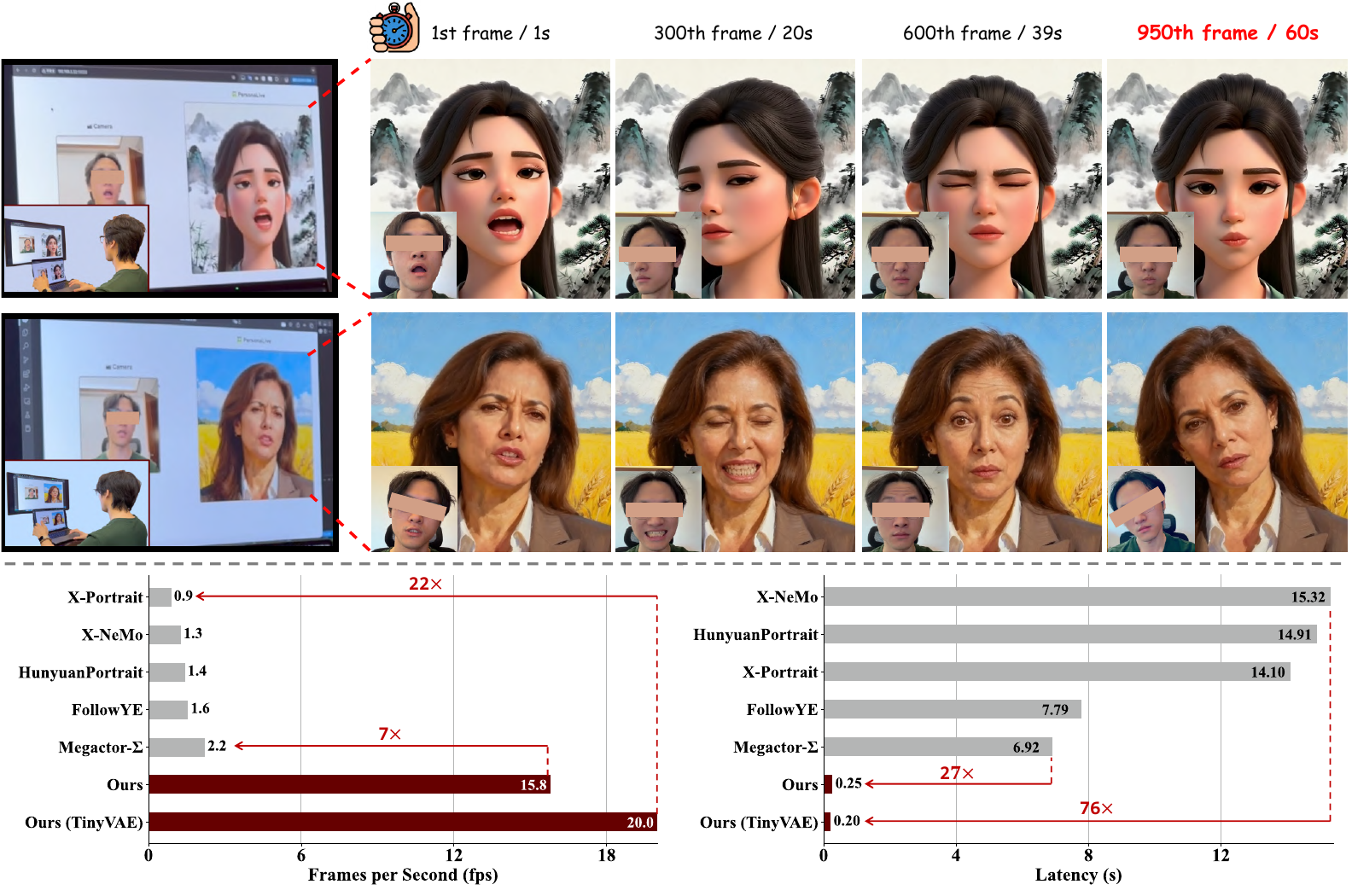}
    \vspace{-2em}
    \captionof{figure}{
    An overview of generated portraits and inference speed of \SHORTNAME. \SHORTNAME{} produces high-quality, temporally stable portrait animations over long sequences, while achieving real-time streaming performance with substantially lower latency than prior diffusion-based approaches.
    }
    \label{fig:teaser}
\end{center}
}]

%% file: 00_abstract.tex
\begin{abstract}
% Abstract goes here.
Current diffusion-based portrait animation models predominantly focus on enhancing visual quality and expression realism, while overlooking generation latency and real-time performance, which restricts their application range in the live streaming scenario. We propose \SHORTNAME, a novel diffusion-based framework towards streaming real-time portrait animation with multi-stage training recipes. 
Specifically, we first adopt hybrid implicit signals, namely implicit facial representations and 3D implicit keypoints, to achieve expressive image-level motion control. Then, a fewer-step appearance distillation strategy is proposed to eliminate appearance redundancy in the denoising process, greatly improving inference efficiency. Finally, we introduce an autoregressive micro-chunk streaming generation paradigm equipped with a sliding training strategy and a historical keyframe mechanism to enable low-latency and stable long-term video generation.
Extensive experiments demonstrate that \SHORTNAME{} achieves state-of-the-art performance with up to \textbf{7-22$\times$} speedup over prior diffusion-based portrait animation models. 
\end{abstract}

%% file: 01_intro.tex
\section{Introduction}
\label{sec:intro}

% Portrait animation—the task of animating a static portrait image according to the motions (expression, pose and location) captured from a driving video—has gained growing attention owing to its vast potential in video conferencing, live streaming and digital human avatars applications. Based on the technical route, portrait animation methods can be categorized into NeRF-based~\cite{pointavatar, otavatar, lam}, GAN-based~\cite{fomm, facevid2vid, dagan, liveportrait, lia-x}, and the more recent diffusion-based approaches~\cite{fadm, x-portrait, x-nemo, hunyuanportrait}. 
Influencers' live streaming has become one of the hottest areas in short-video social media. The Internet provides us with a chance to disguise ourselves as virtual beings. Early 3D avatar approaches~\cite{talkinggaussian, pointavatar, ava} cannot reenact expressive movements and rely on expensive motion capture devices. In contrast, the portrait animation algorithms~\cite{fadm, x-portrait, x-nemo, hunyuanportrait} animates a static portrait image according to the motions~(\ie, detailed expression, pose) captured from a driving video, which shows great potential.

Recently, diffusion-based portrait animation methods~\cite{fadm, x-portrait, x-nemo, hunyuanportrait} have emerged as a dominant paradigm due to their strong generative capabilities. 
% However, achieving high-quality, real-time streaming portrait animation with diffusion models remains a major challenge.due to its computational cost and limited generated video context
% \xd{
However, directly using these models in a live streaming scenario has two key obstacles:
% }: 
%
\textit{(i) the high computational cost}. Current methods primarily focus on improving visual quality and motion consistency while overlooking inference efficiency. Most of them require over 20 denoising steps~\cite{ddim} and rely on the CFG technique~\cite{cfg} to enhance visual fidelity and expression control, which hinders their practical application; 
\textit{(ii) the limitations of chunk-wise processing}. Due to computational and memory constraints, current methods divide long videos into multiple fixed-length chunks and process them independently. To improve temporal consistency across chunks, several methods~\cite{x-portrait, megactor, x-nemo, hunyuanportrait} introduce \textit{training-free} overlapping frames between adjacent chunks, resulting in redundant computation and increased latency. Other methods~\cite{hallo, emo, loopy} reuse the last few frames from the previously generated chunk to enhance cross-chunk consistency, which inevitably causes error accumulation during long video generation. 

% \xiaodong{Should we add a paragraph to introduce the difficulties when above method applied on live streaming? and introduce our high-level motivation}

% \xd{We argue that the portrait animation task mainly handles the motion changes across similar frames, which might not need several steps. Besides, since chunk-wise generation has the same noise level latent, we consider a frame-wise generation paradigm~\cite{diffusionforcing}, where each frame has different noise levels in the denoising context to reduce the latency of the frames. }
% \xd{
We posit that portrait animation primarily involves modeling motion changes across highly similar frames, a task that may not necessitate extensive denoising steps. Furthermore, in contrast to independent chunk-wise generation, we can directly \textit{train} the model for longer and continuous generation conditioned on previously generated frames' intermediate latents and contexts.
% }

We thus propose \SHORTNAME, a diffusion-based portrait animation framework for real-time, streamable motion-driven animation. 
% \xd{
Building upon the recent success of ReferenceNet-based diffusion animation method~\cite{ x-portrait, x-nemo, followyouremoji}, we incorporate several novel components.
% }
\textit{(i) Motion Transfer with Hybrid Control.} 
% \textcolor{purple}{
For portrait animation, effective motion control is essential to ensure realistic and expressive synthesis. In this work, we adopt hybrid motion signals, composed of implicit facial representations \cite{x-nemo} and 3D implicit keypoints~\cite{facevid2vid, liveportrait}, to achieve simultaneous control of both facial dynamics and head movements. Compared with the 2D landmarks~\cite{animateanyone, magicpose} and motion frames~\cite{x-portrait, megactor-sigma} used in existing methods, 3D implicit keypoints provide a more flexible and controllable representation of head motion.
% }
%
\textit{(ii) Fewer-Step Appearance Distillation.} 
% \textcolor{purple}{
We observe that portrait animation exhibits \textit{appearance redundancy} in the denoising process. Specifically, the structural layout and motion are established in the initial denoising steps, whereas numerous subsequent iterations are inefficiently spent on gradually refining appearance details such as texture and illumination. To address this inefficiency, we introduce an appearance distillation strategy that adapts the pretrained diffusion model to a compact sampling schedule, significantly improving inference efficiency without compromising visual quality.
% }
% Although prior works~\cite{lcm, dmd, dmd2, causvid, selfforcing} have explored diffusion distillation in image and video generation to accelerate inference, little attention has been paid to its application in portrait animation. However, portrait animation involves explicit appearance and motion conditions, making it theoretically well-suited for compressing diffusion processing through distillation to achieve real-time inference. Based on this observation, we introduce a diffusion distillation strategy tailored for portrait animation to accelerate the sampling process. Unlike conventional distillation methods, we leverage the ground-truth frames available in this task as direct supervision and incorporate an adversarial loss~\cite{gan} to refine the generated distribution. 
%
\textit{(iii) Micro-chunk Streaming Video Generation.} 
% \textcolor{purple}{
After accelerating the denoising process with the previous strategy, we further aim to enable low-latency and temporally coherent video generation for real-time streaming applications. In contrast to chunk-wise generation~\cite{animatediff}, which relies on latents with uniform noise levels, we adopt an autoregressive micro-chunk streaming paradigm~\cite{diffusionforcing} that assigns progressively higher noise levels across micro chunks with each denoising window, enabling continuous video generation. To mitigate exposure bias~\cite{exposurebias1, exposurebias2} inherent in the autoregressive paradigm, we design a Sliding Training Strategy (ST) to eliminate the discrepancy between the training and inference stages and an effective Historical Keyframe Mechanism~(HKM) that adaptively selects historical frames as auxiliary references, effectively mitigating error accumulation during streaming generation. 
% }
% Inspired by Diffusion Forcing~\cite{diffusionforcing}, we introduce a streaming processing paradigm that enables continuous and long-range video synthesis. Specifically, with each denoising window, frames are assigned progressively increasing noise levels. This design allows us to emit a clean frame after each single denoising step, naturally eliminating the inter-chunk inconsistencies inherent in conventional chunk-wise processing. To further mitigate exposure bias~\cite{exposurebias1, exposurebias2} and error accumulation, we design a sliding training strategy that optimizes the model on generated frames. \textbf{Historical Keyframe}. When generating unseen content (\ie, regions not present in the reference image), diffusion models may produce subtle appearance variations across frames. In streaming processing, these variations can accumulate and amplify over time, leading to degraded temporal coherence and visual stability. To address this issue, we propose a simple yet effective Historical Keyframe Mechanism (HKM) that adaptively selects historical frames as auxiliary references, effectively enhancing temporal coherence during streaming generation. 
Extensive quantitative and qualitative results show that \SHORTNAME~achieves state-of-the-art performance with up to 7-22$\times$ speedup over prior diffusion-based portrait animation models.

The contributions of this paper can be summarized as:
\begin{itemize}
    \item We propose \SHORTNAME, a few-step diffusion-based framework for real-time, streamable portrait animation that achieves low-latency and stable long-term quality.
    \item We design hybrid motion signals combining implicit facial representations and 3D implicit keypoints to enable the simultaneous control of both fine-grained facial dynamics and head movements. Furthermore, we introduce a fewer-step appearance distillation strategy to eliminate appearance redundancy in denoising, greatly improving inference efficiency without compromising visual fidelity.
    \item We design an autoregressive micro-chunk streaming generation paradigm equipped with a sliding training strategy and a historical keyframe mechanism, effectively mitigating exposure bias and error accumulation for stable long-term generation.
    \item Extensive experiments demonstrate that our method achieves state-of-the-art performance while achieving significantly higher efficiency.
\end{itemize}
% }

% To insert a figure: \input{figs/template}
% Or table: \input{tables/template}

%% file: 02_related.tex
\section{Related Work}
\label{sec:related}
\input{figs/overview}

\noindent\textbf{Diffusion-based Portrait Animation.}
Diffusion models~\cite{ddpm, ddim, sde} have demonstrated strong generative capabilities, with Latent Diffusion Models (LDMs)~\cite{ldm} further improving efficiency by performing the denoising process in a lower-dimensional latent space. Building upon this foundation, several works~\cite{x-portrait, x-nemo, hunyuanportrait, fantasyportrait} extend pre-trained diffusion models~\cite{ldm, svd, wan} to controllable and high-fidelity portrait animation with explicit structural conditions, such as facial keypoints~\cite{face-adapter, followyouremoji, skyreels-a1}, facial mesh renderings~\cite{lcvd, mvportrait}, and original driving video~\cite{megactor, x-portrait, megactor-sigma}. These methods typically employ ControlNet~\cite{controlnet} or PoseGuider~\cite{animateanyone} to incorporate motion constraints into the generation process. To model fine-grained facial dynamics, recent works~\cite{x-nemo, fantasyportrait, hunyuanportrait, dreamactor-m1} introduce implicit facial representations. This strategy enhances the preservation of intricate facial expression details, enabling more flexible and realistic animation. However, the above methods primarily focus on improving visual quality and motion consistency while overlooking inference efficiency. In this work, we address this limitation by introducing a real-time, streamable diffusion framework that enables efficient and temporally coherent portrait animation.

\noindent\textbf{Long-term Portrait Animation.}
With the rapid advancement of animation methods and rising user expectations, producing temporally coherent long-term videos has become critical. Due to computational constraints, existing diffusion-based methods~\cite{x-portrait, megactor, megactor-sigma, x-nemo, hunyuanportrait, followyouremoji, hallo} are trained on short clips and rely on inference-time extension for longer sequences. X-Portrait~\cite{x-portrait} and X-NeMo~\cite{x-nemo} adopt the prompt traveling technique~\cite{edge} to enhance temporal smoothness across chunk boundaries. Follow-your-emoji~\cite{followyouremoji} design a coarse-to-fine progressive strategy that generates intermediate frames through keyframe-guided interpolation. Sonic~\cite{sonic} builds global inter-clip connections through the time-aware shifted windows that bridge the preceding clip along the timesteps axis. Despite these advances, existing approaches remain unsuitable for real-time streaming generation. While several methods~\cite{hallo, emo, loopy} leverage ``motion frames'' to enable chunk-wise streaming generation of long videos, they introduce additional training overhead and inevitable error accumulation~\cite{stableavatar}. In contrast, we introduce an autoregressive micro-chunk framework to enable streaming and temporally coherent long-term portrait animation.

\noindent\textbf{Diffusion Model Acceleration.}
Despite their strong performance, the high computational cost of diffusion models keeps them far from real-time applications. Existing acceleration strategies can be broadly categorized into model quantization~\cite{q-diffusion, mpq-dm, sana} and sampling step reduction~\cite{dmd, dmd2, lcm, causvid, selfforcing}. ADD~\cite{sd-turbo} combines
an adversarial and a score distillation objective to efficiently distill diffusion models. Viewing the guided reverse diffusion process as solving an augmented probability flow ODE (PF-ODE), LCMs~\cite{lcm} directly predict the solution of such ODE in latent space, mitigating the need for numerous iterations. DMD~\cite{dmd} and DMD2~\cite{dmd2} distill a many-step diffusion model into a few-step generator by minimizing the approximate Kullback-Liebler (KL) divergences between the diffused target and generator output distributions. Despite recent advances, little attention has been paid to the application of the distillation technique in portrait animation. In this paper, we explore diffusion distillation for real-time portrait animation.

%% file: figs/overview.tex
\begin{figure*}[t]
    \centering
    \includegraphics[width=\textwidth]{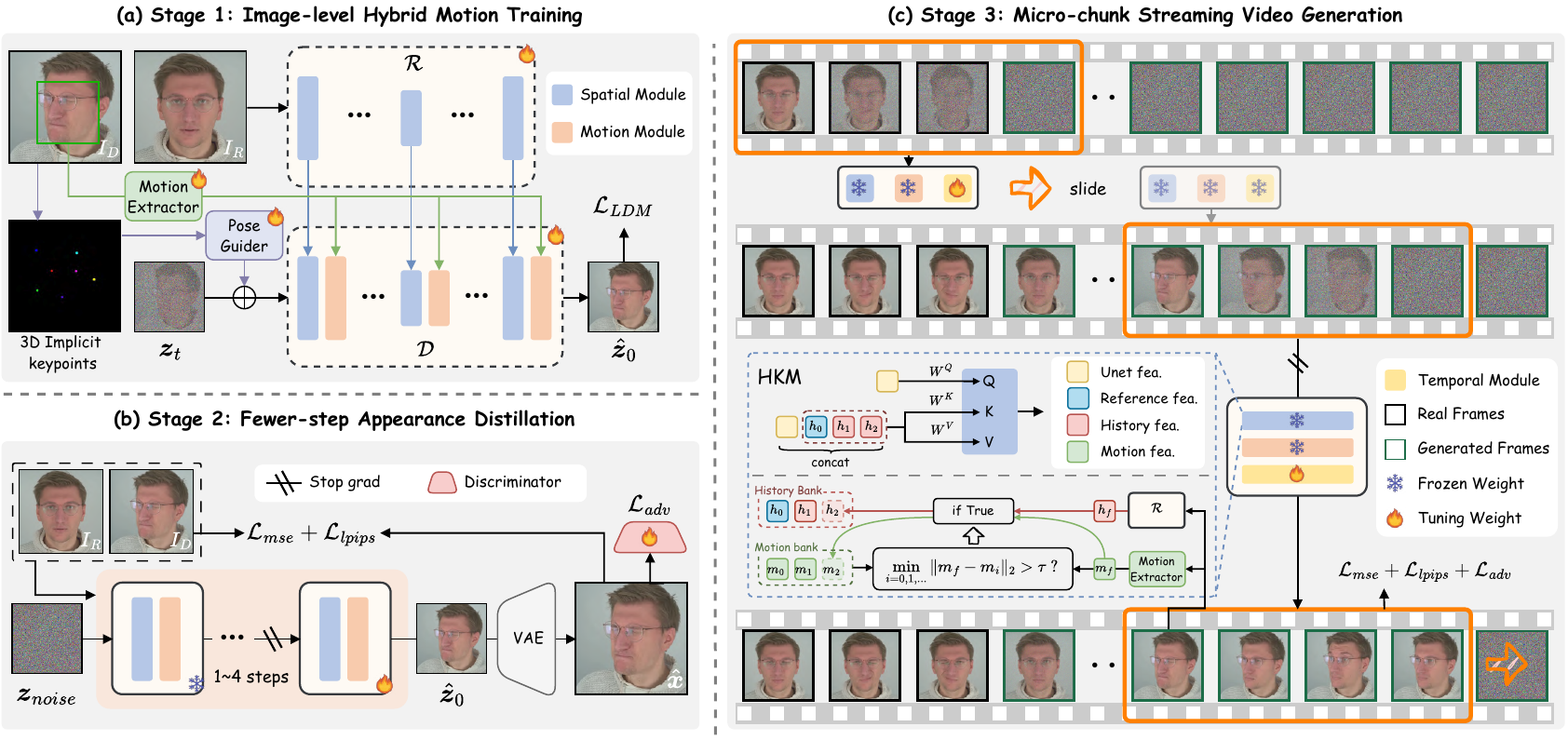}
    \vspace{-2em}
    \caption{Overview of the three-stage pipeline of \SHORTNAME. 
    (a) Image-level hybrid motion training: Learns expressive motion control using implicit facial representations and 3D implicit keypoints. (b) Fewer-step appearance distillation: Eliminates appearance redundancy in the denoising process, improving inference efficiency without compromising visual quality. (c) Micro-chunk streaming video generation: An autoregressive micro-chunk paradigm, equipped with sliding training and historical keyframes, enables low-latency and temporally coherent real-time video generation.
    }
    \vspace{-1em}
    \label{fig:overview}
\end{figure*}

%% file: 03_method.tex
\section{Method}
\label{sec:method}

%（定义任务）
% \xd{
Streaming portrait animation aims to generate long-term, temporally coherent animation streams from a given reference image and driving video, in a real-time and low-latency manner.
% }
%
Formally, given a reference portrait image $I_R$ and a continuous stream of $S$ driving frames $\{I_D^1,I_D^2,\dots,I_D^S\}$, the objective of streaming portrait animation is to synthesize an animation sequence $\mathcal{A}_{\{1,2,...,S\}}$ in a streaming paradigm, where each frame is rendered in real time by combining the appearance information from $I_R$ with the motion cues extracted from $\{I_D^1,I_D^2,\dots,I_D^S\}$, which is formulated as:
\begin{equation}
    \mathcal{A}_i = \mathcal{D}(\mathcal{M}(I_D^i), \mathcal{R}(I_R)),\ i=1,2,\dots,S,
\end{equation}
where $\mathcal{D}$ is the denoising backbone, $\mathcal{M}$ is the motion extractor, and $\mathcal{R}$ is the appearance extractor.
As shown in Fig.~\ref{fig:overview}, we achieve expressive and coherent streaming animation through a three-stage pipeline. We first employ hybrid motion control to achieve expressive and robust motion transfer (Sec.~\ref{sec:motion_transfer}). Then, a fewer-step appearance distillation strategy is introduced to compress the redundant appearance refinement process (Sec.~\ref{sec:distillation}). Finally, to ensure low-latency and stable long-term generation, we propose a micro-chunk streaming generation paradigm equipped with a sliding training strategy and a historical keyframe mechanism (Sec.~\ref{sec:streaming}).

% Specifically, we employ hybrid conditioning signals to enable expressive and robust motion control (Sec.~\ref{sec:motion_transfer}), a few-step distillation strategy to accelerate denoising process (Sec.~\ref{sec:distillation}), and a sliding training strategy coupled with a historical keyframe mechanism to achieve low-latency and temporal coherence generation (Sec.~\ref{sec:streaming}). Below, we first introduce the network architecture.

% \subsection{Preliminary: Latent Diffusion Model}
% Latent Diffusion Models (LDMs)~\cite{ldm} conduct the diffusion process in a compact latent space for improving efficiency. Given an image $x \in \mathbb{R}^{H\times W\times 3}$, an encoder $\mathcal{E}$ first maps it into a latent representation $z = \mathcal{E}(x)$. After that, the latent representation is progressively corrupted by Gaussian noise $\epsilon$ and a UNet~\cite{unet} $\epsilon_{\theta}$ is used to remove added noise from the noisy latent. Formally, $\epsilon_{\theta}$ is trained using the following objective:
% \begin{equation}
%     \mathcal{L}_{LDM}=\underset{t,z,\epsilon}{\mathbb{E}}\Vert \epsilon - \epsilon_\theta(z_t,t,c)\Vert^2_2,
% \end{equation}
% where $z_t$ is the noisy latent of $z$ from timestep $t$ and $c$ represents the conditioning input. In portrait animation, $c$ contains appearance features extracted from the reference portrait and motion features derived from the driving video.

% \subsection{Motion Transfer with Hybrid Control}
\subsection{Image-level Hybrid Motion Training}
\label{sec:motion_transfer}
As shown in Fig.~\ref{fig:overview}~(a), we leverage a pretrained diffusion model $\mathcal{D}$ as the denoising backbone and a reference network $\mathcal{R}$ for appearance conditioning. To achieve expressive and robust motion control, we adopt hybrid conditioning signals composed of implicit facial representations and 3D implicit keypoints. Specifically, we first crop the face region from the driving image $I_D$ and use a face motion extractor $\mathcal{E}_f$~\cite{x-nemo} to encode it into 1D facial motion embeddings $m_f=\mathcal{E}_f(I_D)$, which are then injected into  $\mathcal{D}$ via cross-attention layers. Since the implicit facial representations focus solely on local facial dynamics, we further introduce 3D implicit keypoints to capture global pose, position, and scale information. We use an off-the-shelf method $\mathcal{E}_k$~\cite{liveportrait} to extract 3D parameters from the driving image $I_D$ and the source image $I_R$:
\begin{equation}
    \begin{cases}
        k_{c,d},\ R_d,\ t_d,\ s_d = \mathcal{E}_k(I_D), \\
        k_{c,s},\ R_s,\ t_s,\ s_s = \mathcal{E}_k(I_R),
    \end{cases}
\end{equation}
where $k_{c}$ represents the canonical keypoints, $R$, $t$, and $s$ represent the rotation, translation, and scale parameters, respectively. The driving 3D implicit keypoints $k_d$ are transformed as follows:
\begin{equation}
    k_d = s_d \cdot k_{c,s}R_d + t_d.
\end{equation}
Finally, the extracted 3D implicit keypoints $k_d$ are mapped to the pixel space and injected into $\mathcal{D}$ via PoseGuider~\cite{animateanyone}.

\input{figs/trajectory}

\subsection{Fewer-step Appearance Distillation}
\label{sec:distillation}
%Observation：在portrait animation任务中，motion（结构信息）在第一个去噪就已经基本确定，后续的去噪过程主要是对外观进行细化 ——>外观细化过程中存在大量的冗余 -> 提出Fewer-Step Appearance Distillation strategy.

% \xiaodong{We need to connect the subsection of different stages, here, we should start with our observation, for example, most of the redundancy is the appearance.}

% \textcolor{purple}{
Building upon the hybrid motion control, we observe that in portrait animation, the motion and structural layout of each frame are largely determined during the earliest denoising step, while subsequent iterations primarily refine appearance details, as shown in Fig.~\ref{fig:trajectory}. This observation reveals substantial redundancy in the denoising process, motivating us to develop a distillation strategy that significantly reduces sampling steps without compromising visual fidelity.
% }

Based on the above motivation, we introduce a fewer-step appearance distillation strategy to compress the redundant refinement process into a compact sampling schedule $\{t_i\}_{i=1}^N$, as shown in Fig.~\ref{fig:overview}~(b). Specifically, starting from a Gaussian noise latent $z_{noise}\sim \mathcal{N}(0,I)$, we randomly sample a denoising step $n\in[1,N]$ and perform $n$ denoising iterations to obtain an intermediate noise-free state $\hat{z}_0$, which is then decoded into the pixel space as $\hat{x}=\mathcal{V}_{d}(\hat{z}_0)$. The predicted image $\hat{x}$ is supervised by the corresponding ground-truth frame $x^{gt}$ using a hybrid objective that combines MSE loss, LPIPS loss~\cite{lpips} and adversarial loss~\cite{gan}:
\begin{equation}
    \mathcal{L}_{distill} = \mathcal{L}_{2}(\hat{x},x^{gt}) + \lambda_{lpips}\mathcal{L}_{lpips}(\hat{x},x^{gt}) + \lambda_{adv}\mathcal{L}_{adv}(\hat{x}),
\end{equation}
where $\lambda_{lpips}$ and $\lambda_{adv}$ are balancing coefficients. Backpropagating through the entire diffusion process would result in excessive memory consumption. To improve computational efficiency, we propagate gradients only through the final denoising step, while stochastic step sampling ensures that all middle timesteps receive supervision throughout training.

% \input{tables/distillation}

% \subsection{Stable Streaming Video Generation}
\subsection{Micro-chunk Streaming Video Generation}
\label{sec:streaming}
% While few-step distillation greatly accelerates the denoising process, achieving real-time and temporally coherent video generation remains challenging. A commonly adopted strategy to enhance temporal consistency is chunk-wise generation~\cite{edge}, where long sequences are divided into short, fixed-length segments with overlapping regions for inference. Although effective in improving short-term coherence, this approach introduces redundant computation and delays frame emission, making it unsuitable for real-time streaming generation. To overcome these limitations, we propose a streaming portrait animation framework that processes frames with progressively higher noise levels at each denoising window $W_t = \{x_{t}, \dots, x_{t+N-1}\}$, as shown in Fig.~\ref{fig:overview}(c).
To extend the image animation model for video generation, we integrate a temporal module~\cite{animatediff} into the denoising backbone $\mathcal{D}$. However, instead of assigning a uniform noise level to all frames within a denoising window as in conventional methods, we divide each denoising window into multiple micro-chunks with progressively higher noise levels, as shown in Fig.~\ref{fig:overview}(c). Formally, the denoising window at step $s$ is defined as a collection of $N$ micro-chunks:
\begin{equation}
W_s = \{C^1_s, C^2_s, \dots, C^N_s\},
\end{equation}
\begin{equation}
C_s^n = \{z_i^{t_n} \vert i = 1, 2, \dots, M\}, \ t_1 < t_2 < \dots < t_N,
\end{equation}
where $C^n_s$ denotes the $n$-th micro-chunk consisting of $M$ frames. After each denoising step, all chunks are shifted to lower noise levels, with the first chunk yielding $M$ clean frames ready for emission. Subsequently, the denoising window slides forward by one chunk, and a new noisy chunk $C_{noise}=\{\epsilon_i\}_{i=1}^M$ is appended at the end, initialized with Gaussian noise. This streaming processing paradigm enables continuous frame generation without overlapping regions, ensuring both temporal coherence and low latency.
Despite its efficiency, streaming generation still suffers from exposure bias~\cite{exposurebias1, exposurebias2} and error accumulation when generating long video sequences. To address this, we design a sliding training strategy and a historical keyframe mechanism to jointly stabilize long-range generation and enhance temporal coherence. Below, we give the details of each method.

\noindent\textbf{Sliding Training Strategy.}
The exposure bias in streaming generation primarily stems from the discrepancy between training and inference: during training, the model learns from inputs derived from ground-truth frames. However, during inference, it must rely on its own generated predictions, which inevitably deviate from the distribution of ground-truth data and lead to accumulated temporal errors. To mitigate this issue, we simulate the streaming generation process during training, forcing the model to encounter and learn from its own prediction errors. As shown in Fig.~\ref{fig:overview}~(c), the first denoising window is constructed from noisy ground-truth frames. For $n=1,2,\dots,N-1$, we define:
\begin{equation}
    C_0^n=\{\sqrt{\bar{\alpha}_{t_n}}z_{i}^{gt}+\sqrt{1-\bar{\alpha}_{t_n}}\epsilon_i\}_{i=1}^{M},
\end{equation}
where $\epsilon_i \sim \mathcal{N}(0,I)$, $\alpha_{t_n}$ is a noise scheduling parameter, and $\bar{\alpha}_{t_n}=\prod_{i=1}^{t_n} \alpha_i$. The final chunk $C_0^N$ is initialized with a random noisy chunk $C_{noise}$. 
After each denoising step, the denoising window slides forward by one chunk, and a new noisy chunk is appended at the end, which is completely consistent with the inference procedure. To reduce computational overhead, we compute gradients for only a subset of denoising windows and propagate them through a single denoising step. The overall training objective remains consistent with the appearance distillation stage. As shown in Fig.~\ref{fig:interpolation}, interpolating the implicit motion signals enables a smooth transition from the source motion to the driving motion. Leveraging this property, we introduce a Motion-Interpolated Initialization~(MII) strategy, which constructs the first denoising window using the reference image $I_R$ combined with interpolated implicit motion signals, to align the inference procedure with the training setup.

\input{figs/interpolation}

\noindent\textbf{Historical Keyframe Mechanism.}
When synthesizing regions not explicitly constrained by the reference image (\eg, occluded areas), the stochasticity inherent in diffusion sampling can introduce subtle appearance variations across frames. In a streaming generation setting, these inconsistencies may gradually accumulate, leading to temporal drift and degraded visual stability over time. To mitigate this, we introduce historical keyframes, \ie, representative frames from previously generated results, as auxiliary references, providing the model with stable historical cues to preserve appearance consistency during long-term streaming synthesis. 
As shown in Fig.~\ref{fig:overview}(c), we maintain a history bank $\mathcal{B}_{his}$ and a motion bank $\mathcal{B}_{mot}$. The history bank stores reference features $\{h_0, h_1, \dots\}$ extracted from historical keyframes, while the motion bank stores their corresponding motion embeddings $\{m_0, m_1, \dots\}$. After each denoising step, given the current motion embedding $m_f$ of the first frame, we measure its similarity to $\mathcal{B}_{mot}$ as:
\begin{equation}
    d = \underset{i=0,1,...}{\min}\Vert m_f - m_i\Vert_2.
\end{equation}
If $d > \tau$, where $\tau$ denotes a predefined motion threshold, the current frame is identified as a keyframe. Its reference features $h_f$ and motion embedding $m_f$ are then added to $\mathcal{B}_{his}$ and $\mathcal{B}_{mot}$, respectively. During subsequent inference, these selected historical features are concatenated with the source image feature $h_0$ and injected into the diffusion backbone via the spatial module to enhance temporal consistency.

% Overall, based on the proposed three-stage efforts, the appearance of our method can still achieve the state-of-the-art performance with impressive speed in a real-time streaming manner with expressive animation, enabling more real-world applications.

%% file: figs/trajectory.tex
\begin{figure}[t]
    \centering
    \includegraphics[width=\linewidth]{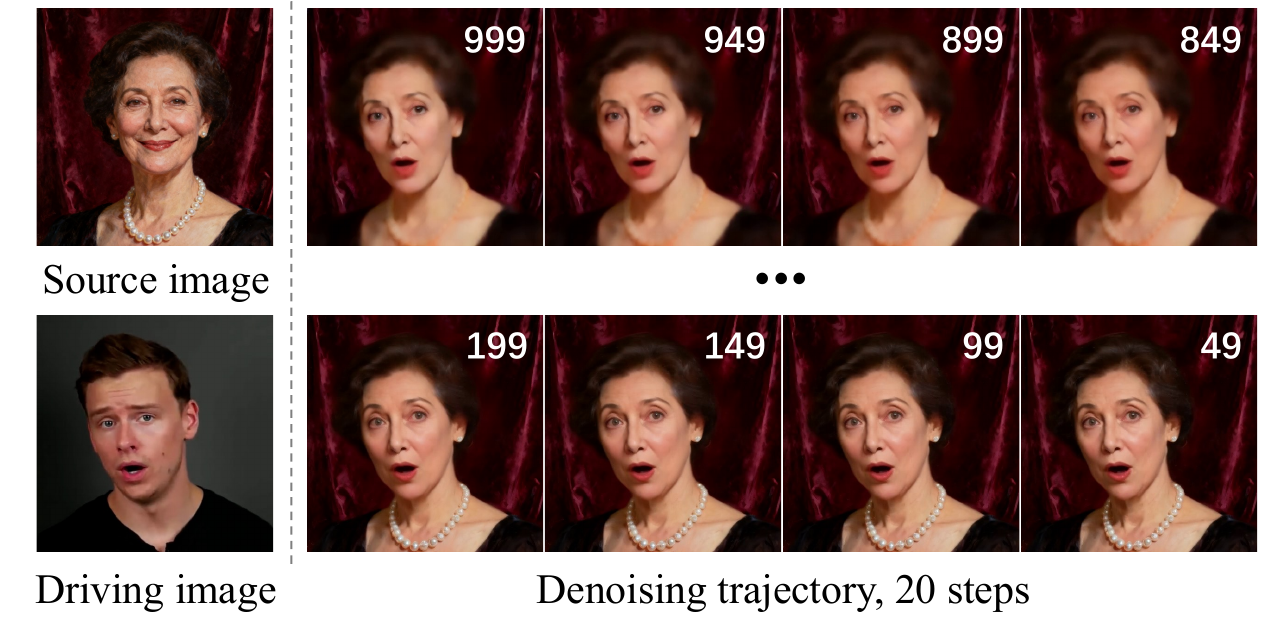}
    \vspace{-2em}
    \caption{The denoising trajectory without CFG~\cite{cfg}. 
    % The motion and structural layout are largely determined at the initial step $t=999$, while subsequent iterations primarily refine appearance details. Adjacent steps exhibit minimal visual change, revealing significant redundancy in the appearance refinement process.}
    }
    \label{fig:trajectory}
\end{figure}

%% file: figs/interpolation.tex
\begin{figure}[t]
    \centering
    \includegraphics[width=\linewidth]{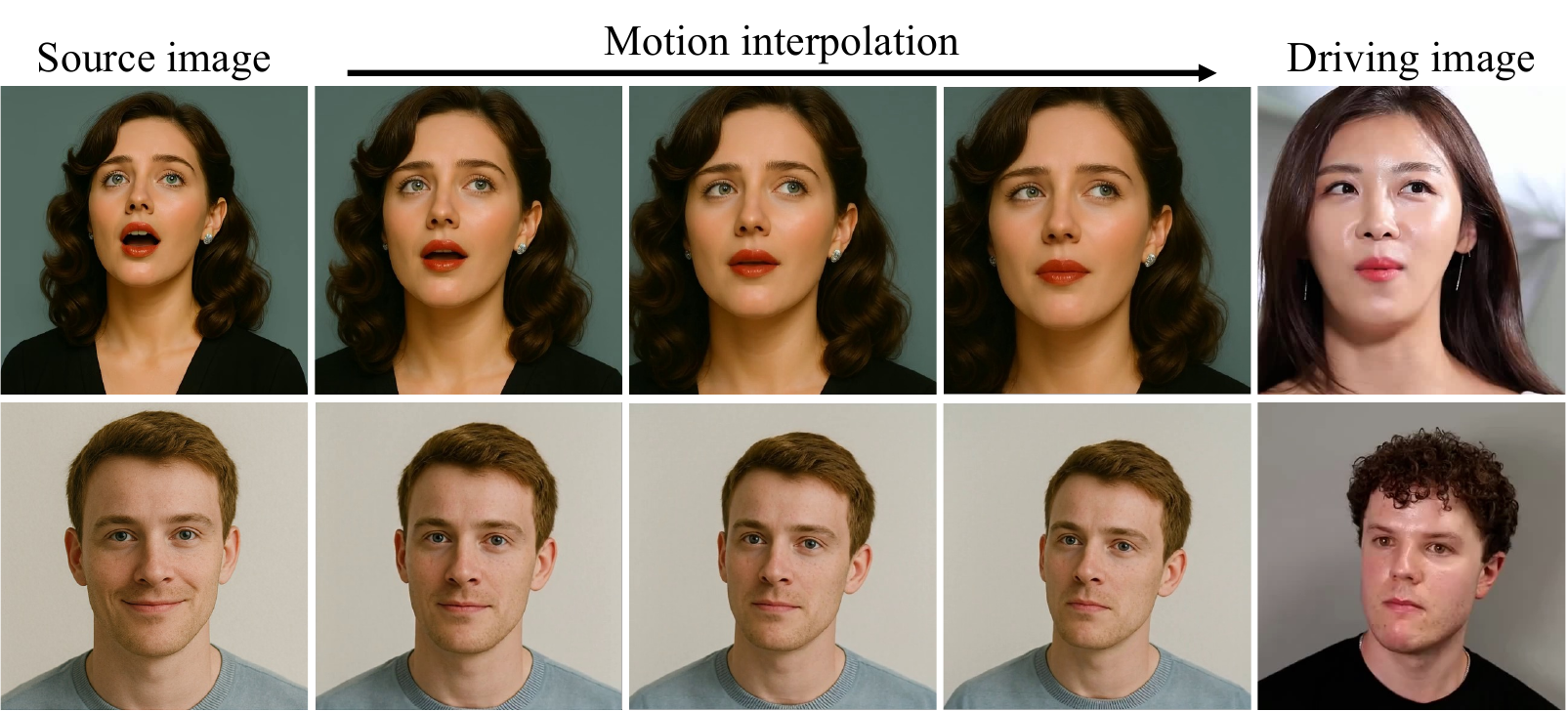}
    \vspace{-1em}
    \caption{Motion interpolation for the first denoising window initialization.
    % Interpolating the motion embeddings and 3D parameters $(R, t, s)$ between the source and driving images enables smooth and natural transitions.}
    }
    \label{fig:interpolation}
\end{figure}

%% file: 04_experiments.tex
\section{Experiments}
\label{sec:experiments}

% \xiaodong{since our method is an efficient streaming method, we need to compare the efficiency and longer generation as main metrics? }

% \subsection{Implementation Details}
We train our method on the VFHQ~\cite{vfhq}, NerSemble~\cite{nersemble} and DH-FaceVid-1K~\cite{dh-facevid} datasets. All data are uniformly processed at 25 fps and cropped to a $512 \times 512$ resolution. For the discriminator, we employ the StyleGAN2~\cite{stylegan2} architecture, initialized with weights pretrained on the FFHQ~\cite{stylegan} dataset. The denoising steps in stage 2 and 3 are set to $N=4$. The chunk size in stage 3 is set to $M=4$. The motion threshold in HKM is set to $\tau=17$.
The training is conducted on 8 Nvidia H100 GPUs using the AdamW optimizer with a learning rate of $1\times10^{-5}$ and a weight decay of 0.01.
% The first training stage is conducted for 30K steps with a batch size of 32. In the second training stage, we apply diffusion distillation to accelerate sampling, reducing the denoising steps to 4 and training for 30K iterations with a batch size of 32. In the third training stage, we train the temporal modules for 30K steps with 40-frames video sequences and a batch size of 8. 
%
Following \cite{liveportrait}, we evaluate our model on the official test split of the TalkingHead-1KH dataset~\cite{facevid2vid}. To further assess performance on long-term portrait animation, we build a benchmark comprising 100 in-the-wild reference portraits and 100 unseen long videos~(most of them longer than one minute), referred to as LV100. More details about implementations can be found in the supplementary materials.

\input{figs/qualitative}
\subsection{Evaluations and Comparisons}

\noindent\textbf{Baselines and Metrics.} We compare our method against state-of-the-art video-driven portrait animation baselines, including the GAN-based LivePortrait~\cite{liveportrait} and Diffusion-based X-Portrait~\cite{x-portrait}, Follow-your-Emoji~\cite{followyouremoji}, Megactor-$\Sigma$~\cite{megactor-sigma}, X-NeMo~\cite{x-nemo}, and HunyuanPortrait~\cite{hunyuanportrait}. RAIN~\cite{rain} adopts the diffusion forcing framework~\cite{diffusionforcing} for streaming generation on anime portrait data. However, it does not address essential challenges such as exposure bias and error accumulation, and the anime portrait domain is overly simplified for real-world portrait animation. Thus, we exclude RAIN from our comparisons.
For self-reenactment, experiments are conducted on the TalkingHead-1KH dataset~\cite{facevid2vid}. We evaluate the performance by computing L1, structural~(SSIM~\cite{ssim}), perceptual~(LPIPS~\cite{lpips}), and temporal~(tLP~\cite{tlp}) differences to assess image quality, motion accuracy, and temporal consistency, respectively. 
For cross-reenactment, we evaluate on our collected LV100 benchmark, which contains diverse identities and long video sequences. We utilize the ArcFace Score~\cite{arcface} as the identity similarity (ID-SIM) metric. Motion accuracy is calculated as the average L1 distance between extracted expression (AED~\cite{fomm}) and pose parameters (APD~\cite{fomm}) of the generated and driving images using SMIRK~\cite{smirk}, with lower values indicating better expression and pose similarity. FVD~\cite{fvd} and tLP~\cite{tlp} are used to evaluate temporal coherence. Furthermore, we report Frames Per Second~(FPS) and the average inter-chunk latency to assess the efficiency of diffusion-based models.

\input{tables/comparison_1}

\noindent\textbf{Self-Reenactment.} For each test video, the first frame is used as the reference image, and the remaining frames serve as the driving inputs and ground-truth targets for sequence generation. As shown in Table~\ref{tab:quant_rec}, despite using significantly fewer denoising steps, \SHORTNAME{} achieves competitive or superior performance across all reconstruction metrics. 
% Benefiting from our three-stage pipeline, our method attains the lowest L1 and LPIPS errors, indicating accurate identity preservation and high perceptual fidelity. Meanwhile, the lower tLP score indicates that \SHORTNAME{} produces smoother and more temporally stable animations compared to prior diffusion-based methods.

\noindent\textbf{Cross-Reenactment.} As evidenced in our qualitative comparisons in Fig.~\ref{fig:qualitative}, \SHORTNAME{} achieves competitive or superior visual fidelity compared to existing methods. It consistently reconstructs facial details and maintains temporal stability across long sequences, while other baselines may exhibit texture smoothing, identity drift, or motion inconsistency in challenging cases. Quantitatively, as reported in Table~\ref{tab:quant_rec}, \SHORTNAME{} achieves performance comparable to existing methods in identity preservation (ID-SIM) and accurate motion transfer (AED/APD), while achieving the best FVD and tLP scores. These results indicate that \SHORTNAME{} provides improved long-term temporal coherence and superior overall perceptual quality.

\noindent\textbf{Efficiency.} As shown in Table~\ref{tab:quant_rec}, the proposed method achieves a substantial improvement in inference efficiency, running at 15.82 FPS with an average latency of only 0.253 s, far surpassing existing diffusion-based baselines. Moreover, by replacing the standard VAE decoder with the TinyVAE~\cite{TinyVAE} decoder, \SHORTNAME{} can further boost the inference speed to 20 FPS. For all diffusion-based competitors, latency is reported without using overlapping frames between chunks. Although this setting allows them to perform chunk-wise streaming generation, the lack of overlapping regions inevitably leads to weaker temporal consistency across chunks.  In contrast, \SHORTNAME{} maintains both real-time performance and stable long-term temporal coherence.

\subsection{Ablation Studies}
To validate the effectiveness of our key components, we conduct comprehensive ablation studies on both the fewer-step appearance distillation strategy and micro-chunk streaming generation paradigm. 

\input{figs/distillation}
\noindent\textbf{Appearance Distillation.} As shown in Fig.~\ref{fig:distillation}, directly reducing the number of sampling steps without distillation (\textit{w/o distill}) leads to significant degradation in visual quality. Incorporating the appearance distillation strategy (\textit{w/ distill, w/o GAN}) effectively improves reconstruction quality; however, the outputs still lack high-frequency details and appear overly smooth. Although applying CFG can enhance fidelity, it substantially reduces inference speed (only 9.5 FPS). In contrast, introducing an adversarial loss enables the model to generate more realistic results without relying on CFG, achieving both high visual fidelity and efficient inference.

\input{tables/ablation}
\input{figs/ablation_stage3}
\noindent\textbf{Micro-chunk Streaming Generation.} To assess the contribution of our streaming design, we examine how each component affects temporal stability and long-range consistency, as shown in Table~\ref{tab:ablation} and Fig.~\ref{fig:stage3}. Below, we give the detailed introduction:

(1) \textit{Sliding Training Strategy}. Removing the sliding training strategy (\textit{w/o ST}) causes the model to train only on GT-constructed noisy inputs, leading to a train-inference mismatch. Since the model never learns to correct its own prediction drift, errors rapidly accumulate and produce severe temporal collapse, as reflected by the huge drop in ID-SIM (0.549) and other metrics. Visual artifacts in Fig.~\ref{fig:stage3}~(last row) clearly show temporal degradation. 

(2) \textit{Historical Keyframe}. As shown in Fig.~\ref{fig:stage3} (\textit{w/o HKM}), removing the historical keyframe mechanism leads to noticeable temporal drift in regions not constrained by the reference portrait (\eg, the clothing area). These inconsistencies accumulate over long sequences, ultimately reducing temporal stability. In contrast, incorporating historical keyframes (highlighted in the yellow box) effectively suppresses such drift and stabilizes long-term generation. Although ID-SIM exhibits a slight decrease, since historical cues partially weaken the reliance on the reference portrait, this trade-off is acceptable given the substantial improvement in temporal coherence.

(3) \textit{Motion-Interpolation Initialization.} To isolate the effect of motion-interpolation initialization, we remove it and instead adopt the variable-length initialization strategy~\cite{pa}. As shown in Fig.~\ref{fig:stage3}~(\textit{w/o MII}), removing MII introduces noticeable appearance distortions at the beginning of the sequence. These artifacts arise from the mismatch between training and inference, as the model is forced to transition abruptly from the reference motion to the driving motion.

(4) \textit{Chunk Size and Attention}. We further examine the influence of micro-chunk structure. Reducing the chunk size from 4 to 2 (\textit{ChunkSize=2}) slightly improves temporal consistency but noticeably degrades identity similarity. This occurs because a smaller chunk size lowers intra-window variation, which helps stabilize short-term dynamics, but it also narrows the effective temporal receptive field, limiting the model’s ability to maintain identity information across longer sequences. As shown in Fig.~\ref{fig:stage3}, a smaller chunk size leads to more artifacts in later frames. Replacing the bidirectional attention with chunk-wise causal attention (\textit{w/ ChunkAttn}) results in similar motion accuracy but a mild decrease in identity similarity.

%% file: figs/qualitative.tex
\begin{figure*}[t]
    \centering
    \includegraphics[width=\textwidth]{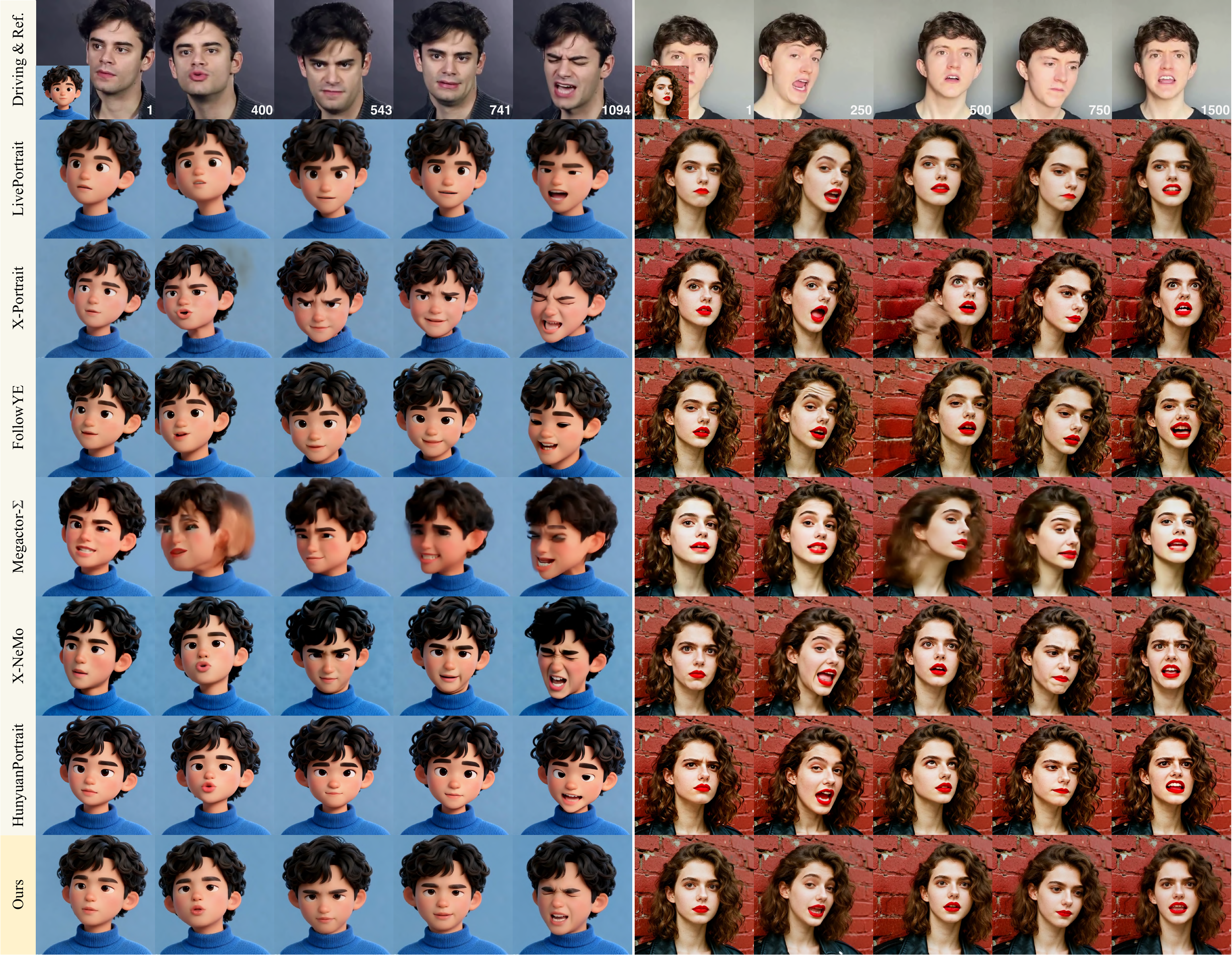}
    \vspace{-1em}
    \caption{Qualitative comparisons. \SHORTNAME{} achieves high-quality portrait animation using significantly fewer denoising steps, while preserving identity, expression fidelity, and facial detail.
    }
    \label{fig:qualitative}
\end{figure*}

%% file: tables/comparison_1.tex
\begin{table*}[t]
\centering
\caption{Quantitative comparisons. Numbers in \textcolor{myred}{\textbf{red}} and \textcolor{myblue}{\textbf{blue}} indicate the best and the second-best results, respectively. tLP multiplied by $10^{-3}$. All speed measurements are conducted on a single NVIDIA H100 GPU. * LivePortrait~\cite{liveportrait} is a frame-wise method using GAN. While it runs significantly faster than diffusion-based approaches, its generated portraits often lack fine-grained details.}
\resizebox{\textwidth}{!}{
\begin{tabular}{lcccc|ccccc|cc}
\toprule
\multirow{2}{*}{Method}  & \multicolumn{4}{c}{\textbf{Self-Reenactment}} & \multicolumn{5}{c}{\textbf{Cross-Reenactment}} &\multicolumn{2}{c}{\textbf{Efficiency}}\\ 
\cmidrule(lr){2-5} \cmidrule(lr){6-10} \cmidrule(lr){11-12} 
& \textbf{L1}\ $\downarrow$ & \textbf{SSIM}\ $\uparrow$ & \textbf{LPIPS}\ $\downarrow$ & \textbf{tLP}\ $\downarrow$ & \textbf{ID-SIM}\ $\uparrow$ & \textbf{AED}\ $\downarrow$ & \textbf{APD}\ $\downarrow$ & \textbf{FVD}\ $\downarrow$ & \textbf{tLP}\ $\downarrow$ & \textbf{FPS}\ $\uparrow$ & \textbf{Latency}\ $\downarrow$\\
\midrule
LivePortrait*~\cite{liveportrait} &0.043	&\textcolor{myred}{\textbf{0.821}}	&0.137	&\textcolor{myred}{\textbf{20.40}} 
&\textcolor{myblue}{\textbf{0.723}} &0.729 &\textcolor{myblue}{\textbf{0.027}} &\textcolor{myblue}{\textbf{557.2}} &\textcolor{myblue}{\textbf{13.51}} &-- &--\\
X-Portrait~\cite{x-portrait} &0.049	&0.777	&0.173	&25.87 
&0.678 &0.823 &0.061 &587.8 &24.52 &0.851 &14.10\\
FollowYE~\cite{followyouremoji} &0.045	&0.803	&0.144	&26.92
&\textcolor{myred}{\textbf{0.773}} &0.911 &0.043 &696.5 &35.13 &1.558 &7.793\\
Megactor-$\Sigma$~\cite{megactor-sigma} &0.055	&0.766	&0.183 &23.55
&0.606 &0.855 &0.079 &585.3 &28.86 &\textcolor{myblue}{\textbf{2.216}} &\textcolor{myblue}{\textbf{6.918}}\\
X-NeMo~\cite{x-nemo} &0.077	&0.689	&0.267	&25.11 
&0.691 &\textcolor{myred}{\textbf{0.679}} &\textcolor{myred}{\textbf{0.022}} &639.1 &18.10 &1.281 &15.32\\
HunyuanPortrait~\cite{hunyuanportrait} &\textcolor{myblue}{\textbf{0.043}}	&0.801	&\textcolor{myblue}{\textbf{0.137}}	&22.33 
&0.644 &0.804 &0.069 &620.4 &16.84 &1.443 &14.91\\
\midrule
% Ours(TinyVAE) \\
Ours &\textcolor{myred}{\textbf{0.039}}	&\textcolor{myblue}{\textbf{0.807}}	&\textcolor{myred}{\textbf{0.129}}	&\textcolor{myblue}{\textbf{21.31}}
&0.698 &\textcolor{myblue}{\textbf{0.703}} &0.030 &\textcolor{myred}{\textbf{520.6}} &\textcolor{myred}{\textbf{12.83}} &\textcolor{myred}{\textbf{15.82}} &\textcolor{myred}{\textbf{0.253}}\\
\bottomrule
\end{tabular}
}
\vspace{-1em}
\label{tab:quant_rec}
\end{table*}

%% file: figs/distillation.tex
% \begin{figure*}[t]
%     \centering
%     \includegraphics[width=\textwidth]{figs/distillation.pdf}
%     \caption{Visualization of the denoising trajectory without CFG technique\cite{cfg}. (a) The baseline model performs 20 steps for high-quality synthesis, yet intermediate steps exhibit minimal visual change, revealing significant redundancy in the sampling process. (b) Reducing steps without adaptation degrades results. (c) Our distilled model achieves comparable quality with only 4 steps.}
%     \label{fig:distillation}
% \end{figure*}

\begin{figure}[t]
    \centering    \includegraphics[width=\linewidth]{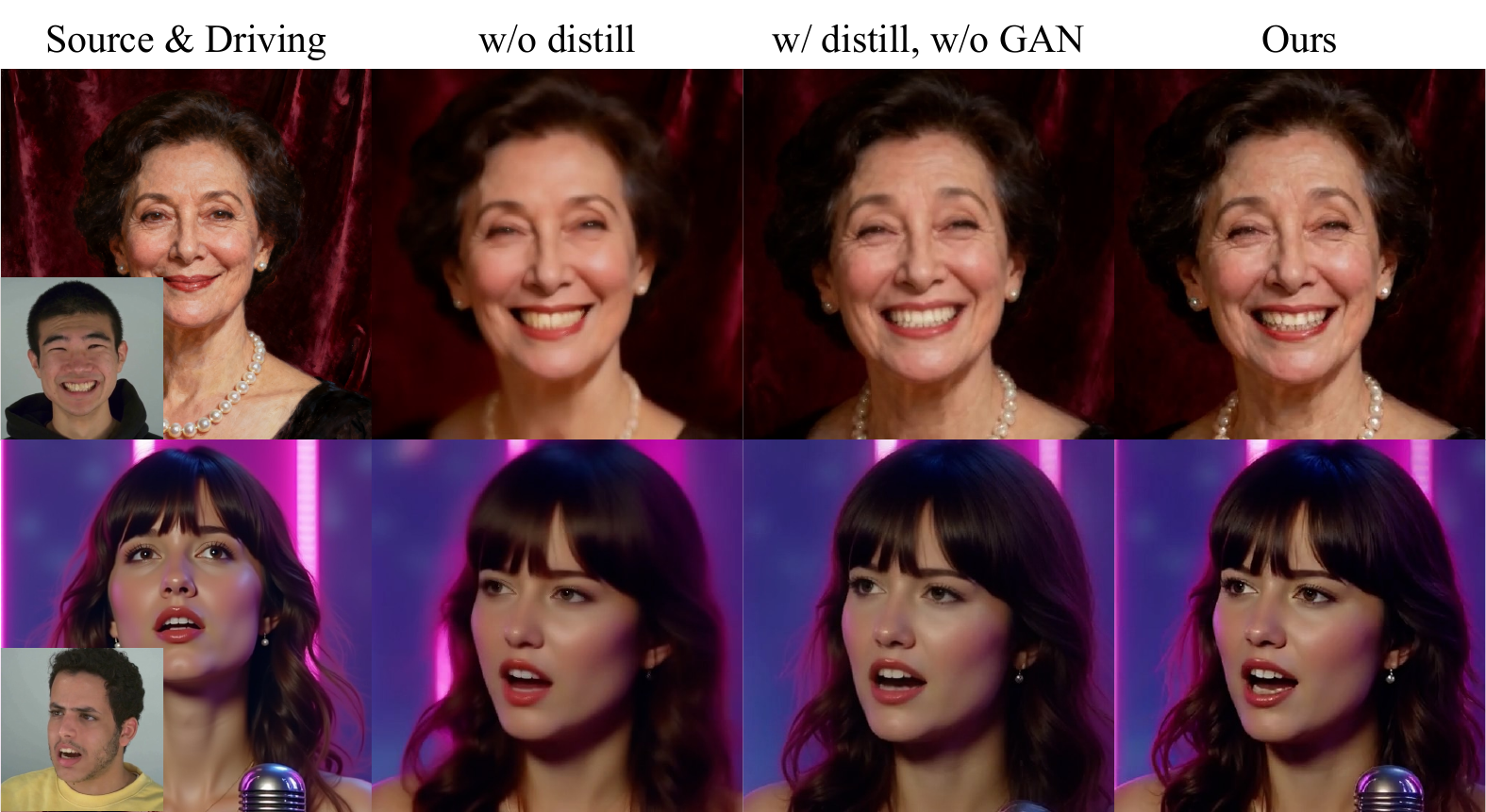}
    \vspace{-1em}
    \caption{Ablation on appearance distillation strategy. All results are generated using 4 denoising steps without the CFG technique.}
    \label{fig:distillation}
\end{figure}

%% file: tables/ablation.tex
\begin{table}[t]
\centering
\caption{Ablation study on micro-chunk streaming generation.}
\resizebox{\linewidth}{!}{
\begin{tabular}{lccccc}
\toprule
setting &\textbf{ID-SIM}$\uparrow$  &\textbf{AED}$\downarrow$ &\textbf{APD}$\downarrow$ &\textbf{FVD}$\downarrow$
&\textbf{tLP}$\downarrow$\\
\midrule
w/ ChunkAttn &0.689 &0.709 &0.032 &537.0 &12.83\\
ChunkSize=2 &0.660 &0.713 & \underline{0.031} & \underline{520.2} & \underline{12.14} \\
w/o MII &0.680 & \textbf{0.703} & \underline{0.031} &\textbf{511.5} &13.06\\
w/o HKM & \textbf{0.728} & 0.710 & \underline{0.031} &535.6 &13.27\\
w/o ST &0.549 &0.785 &0.040 & 678.8 & \textbf{10.05} \\
\midrule
Ours & \underline{0.698} & \textbf{0.703} & \textbf{0.030} & 520.6 & 12.83 \\
\bottomrule
\end{tabular}
}
\vspace{-1em}
\label{tab:ablation}
\end{table}

%% file: figs/ablation_stage3.tex
\begin{figure}[t]
    \centering
    \includegraphics[width=\linewidth]{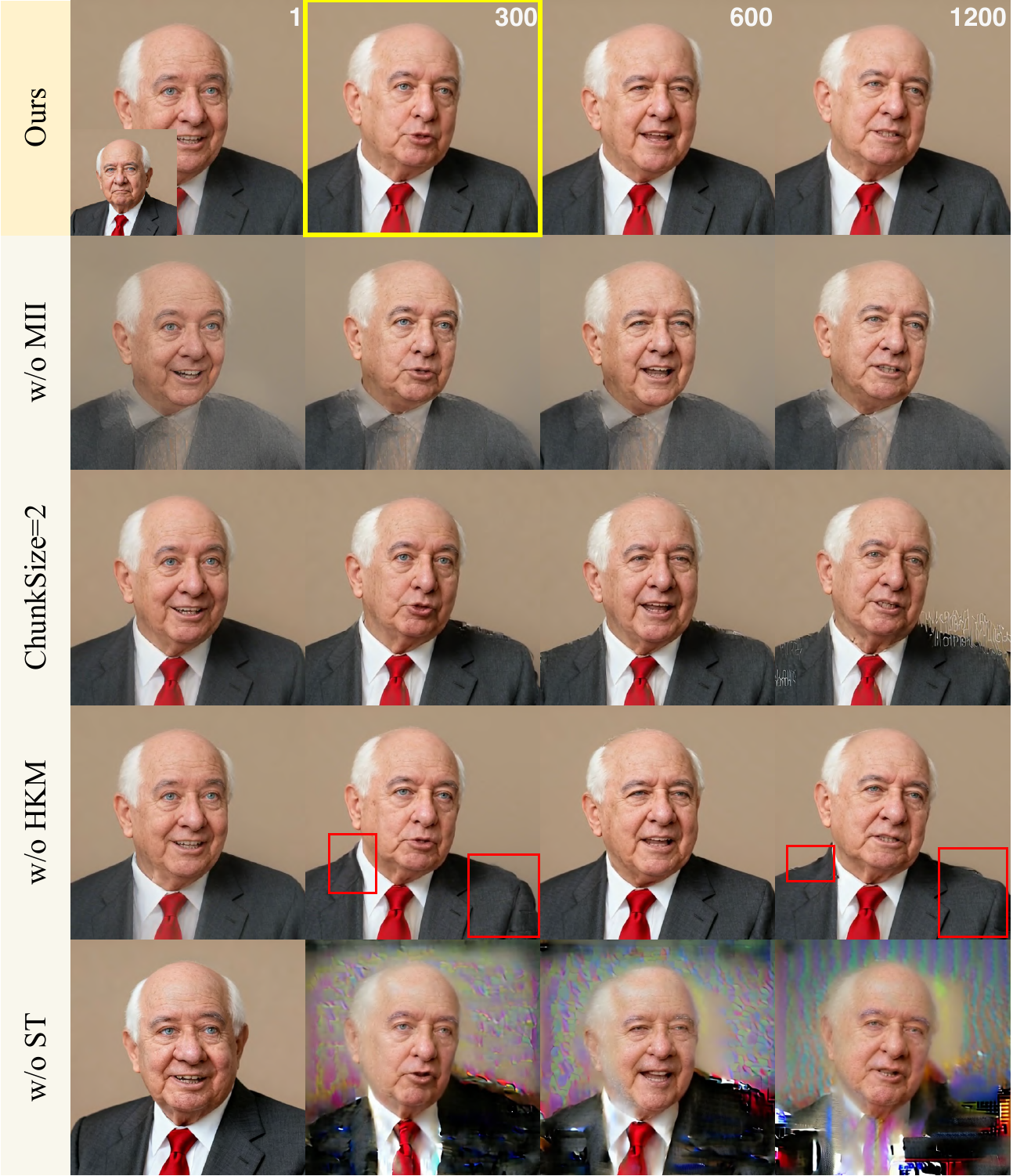}
    \vspace{-1em}
    \caption{Ablation study on the core components of the micro-chunk streaming generation paradigm.
    }
    \label{fig:stage3}
\end{figure}

%% file: 10_conclusion.tex
\input{figs/failure}

\section{Conclusion}
\label{sec:conclusion}
We present \SHORTNAME, an efficient diffusion-based framework for streaming portrait animation via a three-stage strategy. Firstly, we introduce a diffusion-based image animation framework based on hybrid control. Then, by introducing an appearance distillation strategy and a micro-chunk streaming generation paradigm, \SHORTNAME{} enables real-time and low-latency portrait animation. Furthermore, we design a sliding training strategy and a historical keyframe mechanism to alleviate exposure bias and error accumulation, ensuring stable long-term generation and enhanced temporal coherence. We conduct comprehensive experiments to demonstrate the advantages of the proposed methods in terms of visual quality, temporal coherence, and inference efficiency.

\noindent\textbf{Limitation \& Future Work.} While our method achieves real-time and temporally coherent streaming portrait animation, there remain two primary limitations. First, the current framework does not explicitly exploit temporal redundancy across consecutive frames, which could potentially improve inference efficiency and enable longer denoising windows for streaming generation.
Second, our model is trained primarily on human facial data and thus struggles to generalize to out-of-domain portraits with non-human appearances, such as cartoon characters or animals, which may lead to artifacts like blurred or distorted eyes and mouths, as shown in Fig.~\ref{fig:failure}. 
These limitations suggest promising directions for future research in enhancing the scalability and applicability of portrait animation models in real-world scenarios.

% \noindent\textbf{Ethics Statement.}
% Our work focuses on advancing portrait animation technology and is developed solely for academic and creative research. 
% While the method itself is not intended for malicious use, we acknowledge its potential misuse in generating deceptive or non-consensual synthetic media. 
% To promote transparency and responsible use, all generated content should be clearly marked as artificial, and the technology should be applied in accordance with ethical and legal standards.

%% file: figs/failure.tex
\begin{figure}[t]
    \centering
    \includegraphics[width=\linewidth]{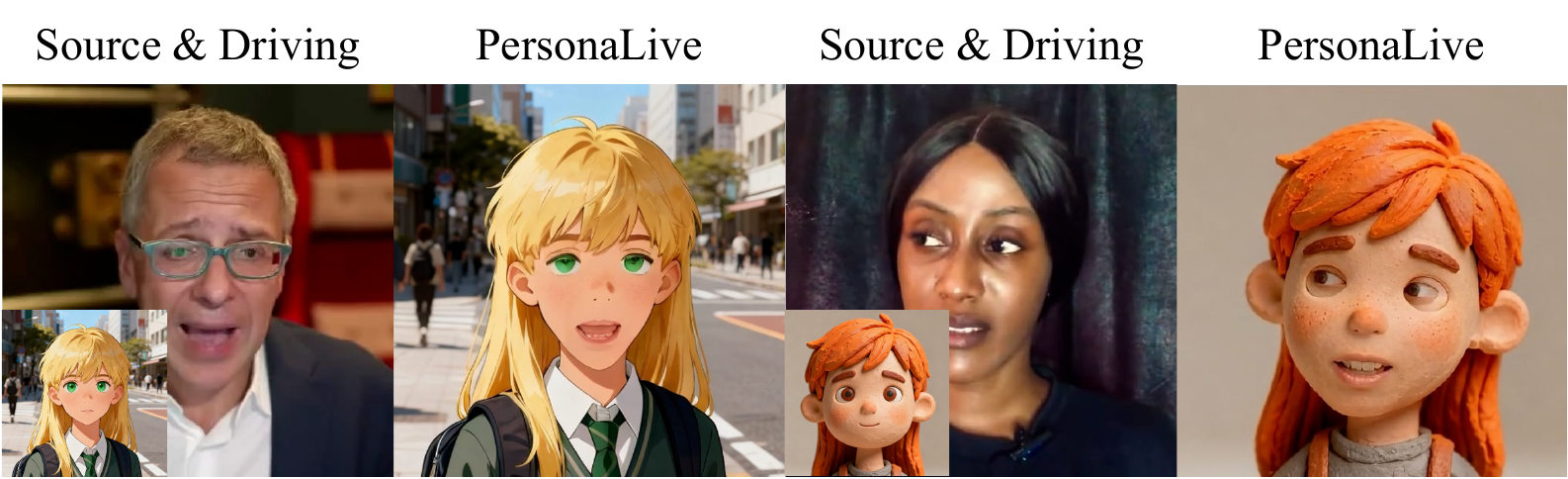}
    \caption{Failure cases. Some details of our method may fail when the given reference images are out of the training domain.}
    \label{fig:failure}
\end{figure}

%% file: 12_appendix.tex
\appendix
\label{sec:appendix}

\section{Preliminary: Latent Diffusion Model}
Latent Diffusion Models (LDMs)~\cite{ldm} conduct the diffusion process in a compact latent space for improving efficiency. Given an image $x \in \mathbb{R}^{H\times W\times 3}$, an encoder $\mathcal{V}_e$ first maps it to a latent representation $z = \mathcal{V}_e(x)$. After that, the latent representation is progressively corrupted by Gaussian noise $\epsilon$ as:
\begin{equation}
    z_t = \Psi(z,\epsilon,t) = \sqrt{\bar{\alpha}_t}z+\sqrt{1-\bar{\alpha}_t}\epsilon,
\end{equation}
where $\bar{\alpha}_t$ is a pre-defined noise schedule within a finite time horizon $t\in[0,1000]$. A U-Net~\cite{unet} denoiser $\epsilon_{\theta}$ is trained to predict and remove the added noise from $z_t$. The training objective is formulated as:
\begin{equation}
    \mathcal{L}_{LDM}=\underset{t,z,\epsilon}{\mathbb{E}}\Vert \epsilon - \epsilon_\theta(z_t,t,c)\Vert^2_2,
\end{equation}
where $c$ represents the conditioning input. In portrait animation, the input is a multi-frame latent window $\{z_t^i\}_{i=1}^M$, $c$ contains appearance features extracted from the reference portrait and motion features derived from the driving video.

\section{Experimental Details}
\noindent\textbf{More implementation details.} 
Our training pipeline progresses through three stages. In Stage~1, we conduct image-level hybrid motion training. Specifically, we randomly sample paired reference and driving images from the training videos, enabling the model to learn appearance conditioning from the reference portrait and motion conditioning from the driving input. This stage is trained for 30K iterations with a batch size of 32. During this stage, all model parameters are updated. 
After this initialization, Stage~2 performs fewer-step appearance distillation following Algorithm \ref{alg:distill}. We adopt a compact sampling schedule $[0, 333, 666, 999]$, enabling the model to learn to reconstruct high-quality frames from only a few denoising steps. This stage is trained for 30K iterations with a batch size of 32, and all model parameters remain trainable.
Stage~3 focuses on temporal modeling: we train the temporal attention layers following Algorithm \ref{alg:slidingtraining}. At each iteration, we slide over a 40-frame sequence and perform three model updates. Only the temporal attention layers are trainable in this stage, while all other parameters remain frozen. This stage is trained for 10K iterations with a batch size of 8. For Stage 2 and 3, $\lambda_{lpips}$ and $\lambda_{adv}$ are set to 2.0 and 0.05, respectively.
\input{tables/distillation}
\input{tables/slidingtraining}

\noindent\textbf{Details of LV100.} 
For long-term portrait animation evaluation, we collect 100 unseen videos ($\geq$1 minute, 25 FPS) from various online platforms, including YouTube, TikTok, and BiliBili. Additionally, we compile 100 in-the-wild reference portraits from ChatGPT-5, Doubao, and Pexels, covering a broad range of facial structures, appearances, and styles. Representative examples from the LV100 benchmark are shown in Fig.~\ref{fig:lv100}.

\input{figs/lv100}

\noindent\textbf{Implicit 3D keypoints.} As shown in Fig.~\ref{fig:keypoints}, the implicit 3D keypoint extractor~\cite{liveportrait} produces 21 canonical keypoints (left), from which we select a subset of stable landmarks (right) to encode global head pose, scale, and spatial configuration. These selected keypoints serve as an effective global motion prior in our hybrid motion control.

\input{figs/keypoints}

\noindent\textbf{Motion-interpolation initialization.}
Given the reference image $I_R$ and first driving frame $I_D^1$, we construct the first denoising window $W_0 = \{C^1,C^2,\dots C^N\}$ using noisy reference latent $z_{ref}=\mathcal{V}_e(I_R)$:
\begin{equation}
    C^n=\{\sqrt{\bar{\alpha}_{t_n}}z_{ref}+\sqrt{1-\bar{\alpha}_{t_n}}\epsilon_i\}_{i=1}^{M},
\end{equation}
where $\epsilon_i \sim \mathcal{N}(0,I)$. Subsequently, we interpolate the motion signals between the reference image and the first driving frame. Let $m_{f,s}$ and $m_{f,d}^1$ denote the implicit facial motion embeddings extracted from $I_R$ and $I_D^1$, respectively. For the $i$-th frame in the initial window:
\begin{equation}
    m_{f,i} = (1-\omega_i) m_{f,s} + \omega_im_{f,d}^1,
\end{equation}
where $\omega_i = \frac{i-1}{MN-1}$ is the interpolation factor. For implicit 3D keypoints, we interpolate the 3D transformation parameters:
\begin{equation}
\begin{aligned}
    R_i &= R\big((1-\omega_i)\theta_s + \omega_i\theta_d^1\big),\\
    s_i &= (1-\omega_i)s_s + \omega_is_d^1,\\
    t_i &= (1-\omega_i)t_s + \omega_it_d^1,
\end{aligned}
\end{equation}
where $\theta = (\text{pitch}, \text{yaw},\text{roll})$ denotes Euler angles and $R(\theta)$ denotes the rotation matrix constructed from Euler angles. The interpolated keypoints are then computed as:
\begin{equation}
    k_i = s_i\cdot k_{c,s} R_i + t_i.
\end{equation}

\section{More Ablations}
In this section, we provide additional ablations on some network and hyperparameters.

\input{figs/motion}
\noindent\textbf{Hybrid motion signals.} As shown in Fig.~\ref{fig:motion}, the implicit 3D keypoints $k_d$ control global head movements, including rotation, translation, and scale. In contrast, the implicit facial motion embedding $m_{f,d}$ primarily controls fine-grained facial expressions. Although $m_{f,d}$ contains some pose-related cues, these signals have lower priority compared to the implicit 3D keypoints, as reflected in the result of $k_s + m_{f,d}$.

\noindent\textbf{Motion threshold.} We evaluate the effect of the motion threshold $\tau$ in our historical keyframe mechanism. As shown in Table~\ref{tab:threshold}, a smaller $\tau$ triggers more frequent history bank updates, providing richer historical information that helps stabilize long-term temporal consistency (lower FVD and tLP). However, more historical frames weakens the influence of the reference image $I_R$, leading to slight ID drift and consequently lower ID-SIM. Conversely, a larger $\tau$ better preserves identity but slightly degrades temporal stability due to fewer historical keyframes. Overall, we set $\tau = 17$ as it offers the best trade-off between identity preservation and temporal coherence.

\noindent\textbf{VAE decoder.} We further analyze the impact of different VAE decoders. As shown in Table~\ref{tab:decoder}, TinyVAE~\cite{TinyVAE} significantly accelerates inference (up to 20 FPS) but introduces noticeable degradation in visual quality. SVD VAE~\cite{svd} provides no improvement in temporal consistency and even reduces runtime efficiency. In contrast, SD VAE~\cite{ldm} achieves the best overall performance while maintaining competitive inference speed.

\input{tables/threshold}
\input{tables/decoder}

\section{More results}
As shown in Fig.~\ref{fig:more_same_1}, Fig.~\ref{fig:more_same_2}, Fig.~\ref{fig:more_cross_1}, and Fig.~\ref{fig:more_cross_2}, we present additional visualization results under self-reenactment and cross-reenactment setting, further demonstrating the robustness and generalization ability of \SHORTNAME. Fig.~\ref{fig:long_video_1}, Fig.~\ref{fig:long_video_2}, Fig.~\ref{fig:long_video_3}, and Fig.~\ref{fig:long_video_4} show long avatar videos synthesized by \SHORTNAME, highlighting its stability and consistency over long-term sequences.

\section{Ethics Statement.}
Our work focuses on advancing portrait animation technology and is developed solely for academic and creative research. While the method itself is not intended for malicious use, we acknowledge its potential misuse in generating deceptive or non-consensual synthetic media. To promote transparency and responsible use, all generated content should be clearly marked as artificial, and the technology should be applied in accordance with ethical and legal standards.

\input{figs/more_qualitative}

%% file: tables/distillation.tex
\begin{algorithm}[t]
\caption{Fewer-step Appearance Distillation}
\label{alg:distill}
\KwIn{Reference image $I_R$; Driving image $I_D$; Animation model $G_\theta$; VAE decoder $\mathcal{V}_d$; Sampling schedule $\{t_i\}_{i=1}^N$}
\KwOut{Updated parameters $\theta$}

\BlankLine
\For{each iteration}{
\tcp{Sample initial noisy latent and random step count $n$}

Sample $z_{\text{noise}} \sim \mathcal{N}(0,I)$\\
Sample $n \sim \mathrm{Uniform}(1,N)$\\
Set $z_{t_N} \leftarrow z_{\text{noise}}$\\

\tcp{Perform $n$ denoising steps}

\For{$i = N$ \KwTo $N-n+1$}{
    \eIf{$i > N-n+1$}{
        Disable gradient computation\\
        Set $\hat{z}_0 \leftarrow G_\theta(z_{t_i}; t_i,I_R,I_D)$\\
        Sample $\epsilon \sim \mathcal{N}(0,I)$\\
        Set $z_{t_{i-1}} \leftarrow \Psi(\hat{z}_0,\epsilon,t_{i-1})$ 
        
    }{
        Enable gradient computation\\
        Set $\hat{z}_0 \leftarrow G_\theta(z_{t_i}; t_i,I_R,I_D)$\\
    }
}

\tcp{Decode prediction}
Set $\hat{x} \leftarrow \mathcal{V}_d(\hat{z}_0)$\\
\tcp{Update model}
Update $\theta$ via Distillation loss $\mathcal{L}_{distill}$}
\Return{$\theta$}
\end{algorithm}

%% file: tables/slidingtraining.tex
\begin{algorithm}[t]
\caption{Sliding Training Strategy}
\label{alg:slidingtraining}
\KwIn{
Reference image $I_R$; 
Driving video $\{I_D^i\}_{i=1}^{S}$; 
Animation model $G_\theta$; 
VAE encoder $\mathcal{V}_e$; 
VAE decoder $\mathcal{V}_d$; 
Micro-chunk size $M$; 
Sampling schedule $\{t_i\}_{i=1}^N$}
\KwOut{Updated parameters $\theta$}

\BlankLine
\For{each iteration}{
\tcp{Construct the initial denoising window $W_0$}
Set $z^{1:M(N-1)} \leftarrow \mathcal{V}_e\big(I_D^{1:M(N-1)}\big)$\\
\For{$n=1$ to $N-1$}{
    Sample $\epsilon \sim \mathcal{N}(0,I)$\\ 
    Set $C^n \leftarrow \Psi\big(z^{(n-1)M+1:nM},\epsilon,t_n\big)$
}

Sample $C^N \sim \mathcal{N}(0,I)$ \tcp*{Last chunk is pure noise}
Initialize window $W_0 = \{C^1, C^2, \dots, C^N\}$\\

\tcp{Sliding generation and training}
\For{$s=0$ to $\frac{S}{M}-N$}{
Set $V_s \leftarrow I_D^{sMN+1:(s+1)MN}$ \\
\eIf{$s \bmod (N-1) \neq 0$}{
    Disable gradient computation\\
    Set $\hat{W}_s \leftarrow G_\theta(W_s,t_{1:N},I_R,V_s)$
    }{
    Enable gradient computation\\
    Set $\hat{W}_s \leftarrow G_\theta(W_s,t_{1:N},I_R,V_s)$\\
    \tcp{Decode sequence prediction}
    Set $\hat{x}_{seq} \leftarrow \mathcal{V}_d(\hat{W}_s)$\\
    \tcp{Update model}
    Update $\theta$ via Distillation loss\\
    Disable gradient computation\\
    }
    \tcp{Slide window forward}
    Set $W_{s+1} \leftarrow \{\hat{C}^{s+2},\hat{C}^{s+3},\dots,\hat{C}^{s+N}\}$\\
    Sample $\epsilon_{s+1} \sim \mathcal{N}(0,I)$\\
    Set $W_{s+1} \leftarrow \Psi(W_{s+1}, \epsilon_{s+1},t_{1:N-1})$\\
    Sample $C^{s+N+1} \sim \mathcal{N}(0,I)$\\
    Set $W_{s+1} \leftarrow \{W_{s+1},C^{s+N+1}\}$
}
}
\Return{$\theta$}
\end{algorithm}

%% file: figs/lv100.tex
\begin{figure}[t]
    \centering
    \includegraphics[width=\linewidth]{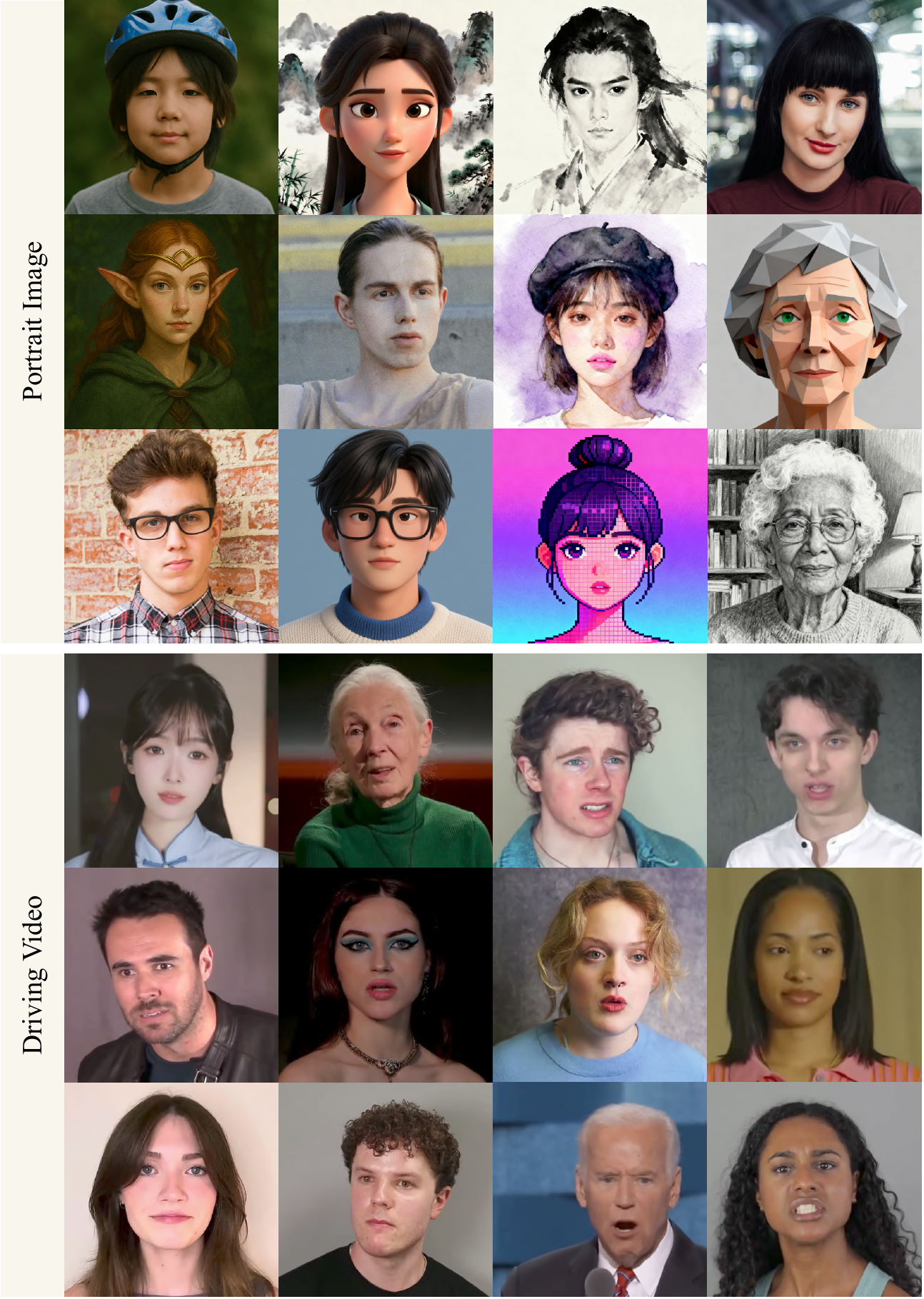}
    \vspace{-2em}
    \caption{Examples from LV100.
    }
    \label{fig:lv100}
\end{figure}

%% file: figs/keypoints.tex
\begin{figure}[t]
    \centering
    \includegraphics[width=\linewidth]{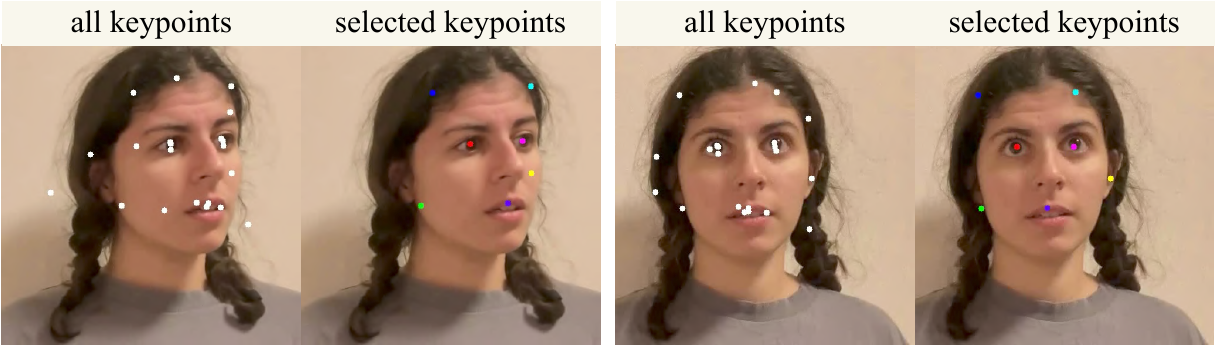}
    \vspace{-2em}
    \caption{The implicit 3D keypoints used in our hybrid motion control.
    }
    \label{fig:keypoints}
\end{figure}

%% file: figs/motion.tex
\begin{figure}[t]
    \centering
    \includegraphics[width=\linewidth]{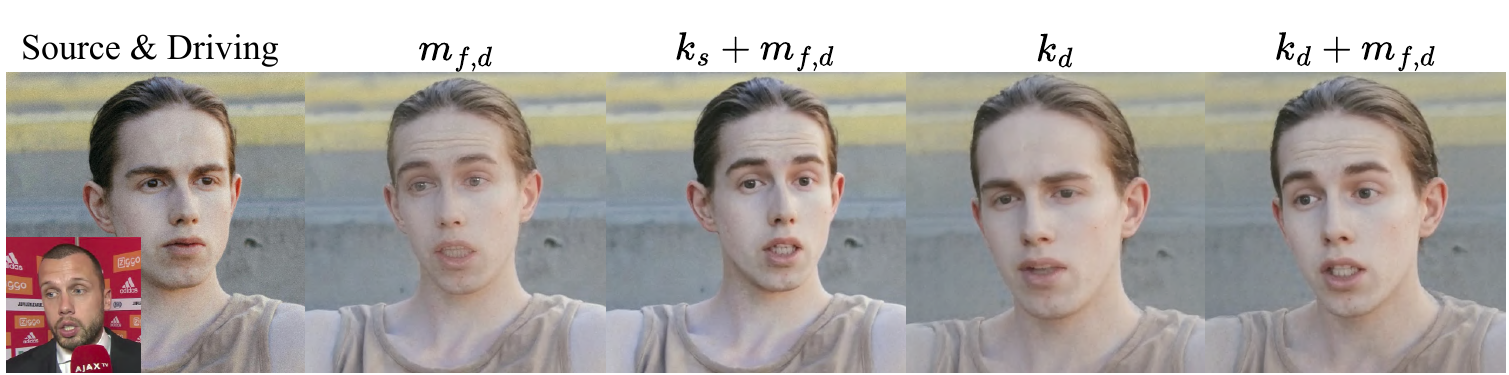}
    \vspace{-2em}
    \caption{Effect of implicit 3D keypoints and facial motion embedding.
    }
    \label{fig:motion}
\end{figure}

%% file: tables/threshold.tex
\begin{table}[t]
\centering
\caption{Ablation study on motion threshold $\tau$.}
\resizebox{\linewidth}{!}{
\begin{tabular}{lccccc}
\toprule
$\tau$ &\textbf{ID-SIM}$\uparrow$  &\textbf{AED}$\downarrow$ &\textbf{APD}$\downarrow$ &\textbf{FVD}$\downarrow$
&\textbf{tLP}$\downarrow$\\
\midrule
15	&0.6924	&0.7043	&0.0309	&510.9	&12.69\\
16	&0.6940	&0.7039	&0.0306	&516.6	&12.78\\
17	&0.6983	&0.7028	&0.0305	&520.6	&12.83\\
18	&0.7015	&0.7047	&0.0306	&522.9	&12.93\\
19	&0.7097	&0.7084	&0.0304	&526.5	&13.05\\
20	&0.7159	&0.7099	&0.0303	&529.2	&13.18\\
\bottomrule
\end{tabular}
}
\vspace{-1em}
\label{tab:threshold}
\end{table}

%% file: tables/decoder.tex
\begin{table}[t]
\centering
\caption{Ablation study on VAE decoder.}
\resizebox{\linewidth}{!}{
\begin{tabular}{lccccc}
\toprule
decoder &\textbf{ID-SIM}$\uparrow$  &\textbf{AED}$\downarrow$ &\textbf{APD}$\downarrow$ &\textbf{tLP}$\downarrow$
&\textbf{FPS}$\uparrow$\\
\midrule
SVD VAE~\cite{svd} &\underline{0.6920}	&\underline{0.7452}	&0.0489		&\underline{12.97}	&11.4\\
SD VAE~\cite{ldm} &\textbf{0.6983}	&\textbf{0.7028}	&\textbf{0.0305}	&\textbf{12.83} &\underline{15.8}\\
TinyVAE~\cite{TinyVAE} &0.6758	&0.7593	&\underline{0.0489}		&14.66	&\textbf{20.0}\\
\bottomrule
\end{tabular}
}
\vspace{-1em}
\label{tab:decoder}
\end{table}

%% file: figs/more_qualitative.tex
\begin{figure*}[t]
    \centering
    \includegraphics[width=\textwidth]{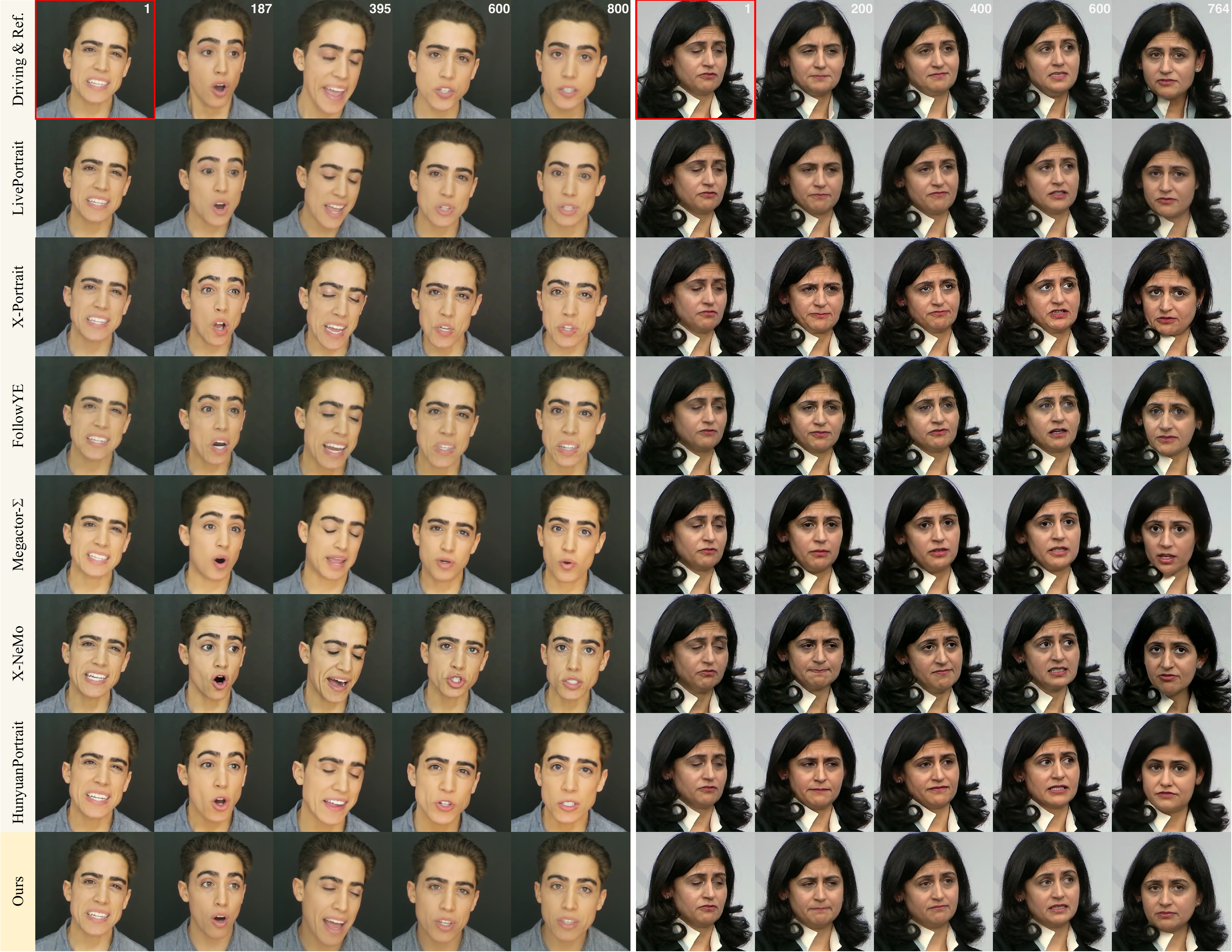}
    \vspace{-1em}
    \caption{More visualizations of self-reenactment comparison (1/2). The images with red borders are the reference images.
    }
    \label{fig:more_same_1}
\end{figure*}

\begin{figure*}[t]
    \centering
    \includegraphics[width=\textwidth]{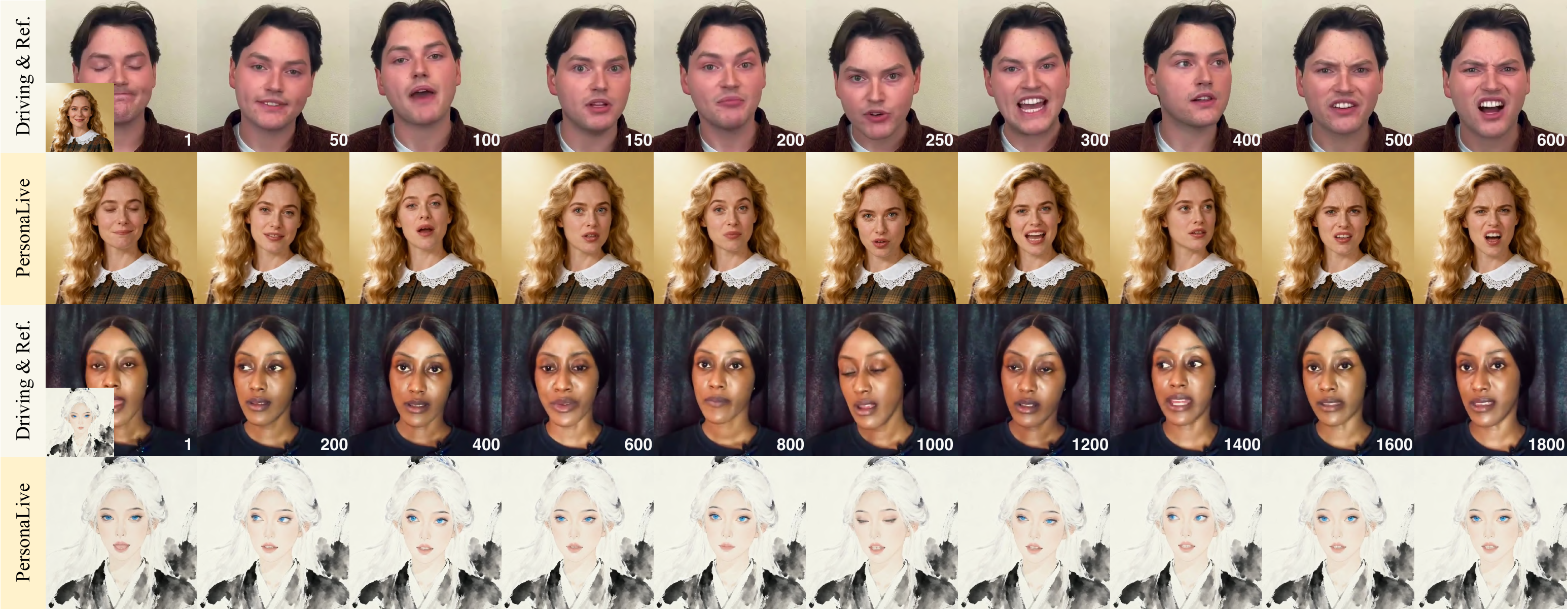}
    \vspace{-1em}
    \caption{Long avatar video results (1/4).
    }
    \label{fig:long_video_1}
\end{figure*}

\begin{figure*}[t]
    \centering
    \includegraphics[width=\textwidth]{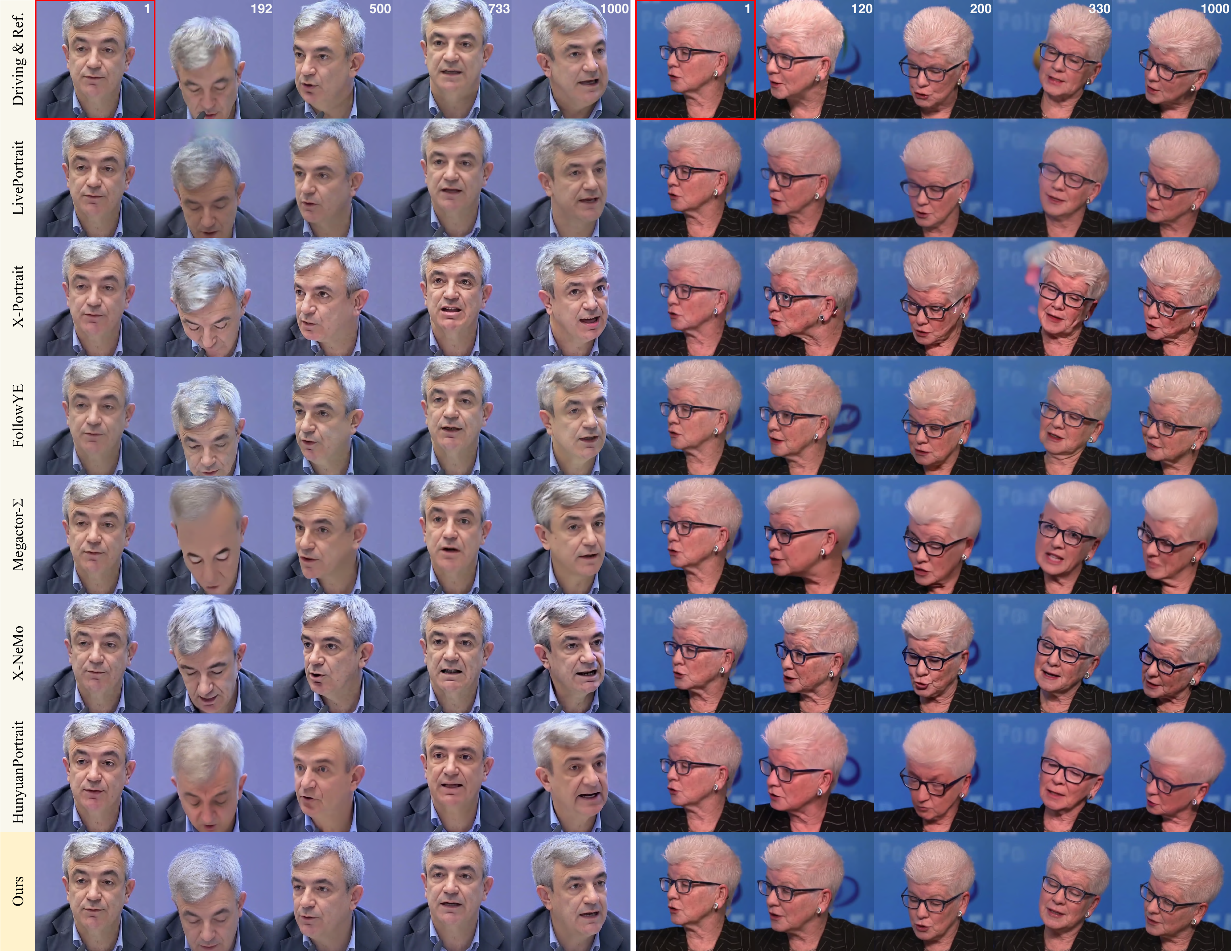}
    \vspace{-1em}
    \caption{More visualizations of self-reenactment comparison (2/2). The images with red borders are the reference images.
    }
    \label{fig:more_same_2}
\end{figure*}

\begin{figure*}[t]
    \centering
    \includegraphics[width=\textwidth]{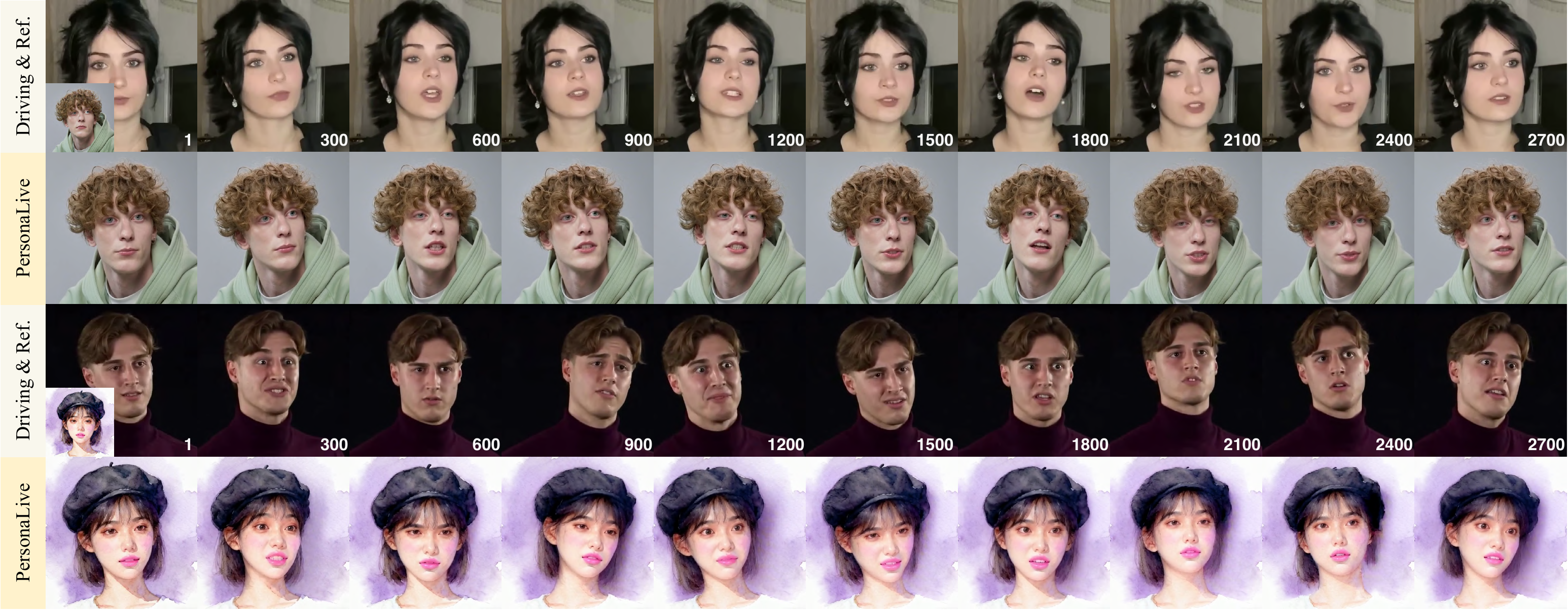}
    \vspace{-1em}
    \caption{Long avatar video results (2/4).
    }
    \label{fig:long_video_2}
\end{figure*}

\begin{figure*}[t]
    \centering
    \includegraphics[width=\textwidth]{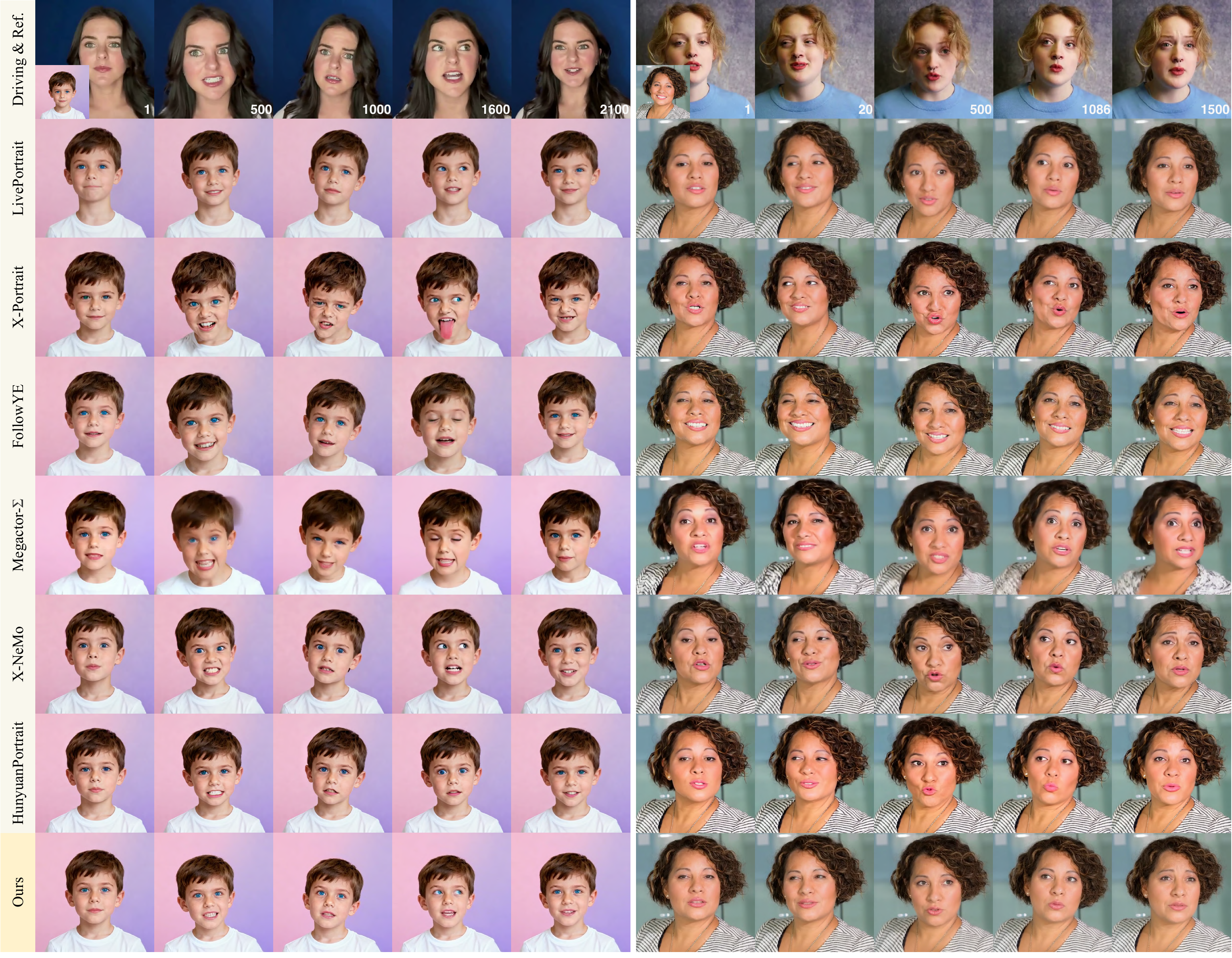}
    \vspace{-1em}
    \caption{More visualizations of cross-reenactment comparison (1/2).
    }
    \label{fig:more_cross_1}
\end{figure*}

\begin{figure*}[t]
    \centering
    \includegraphics[width=\textwidth]{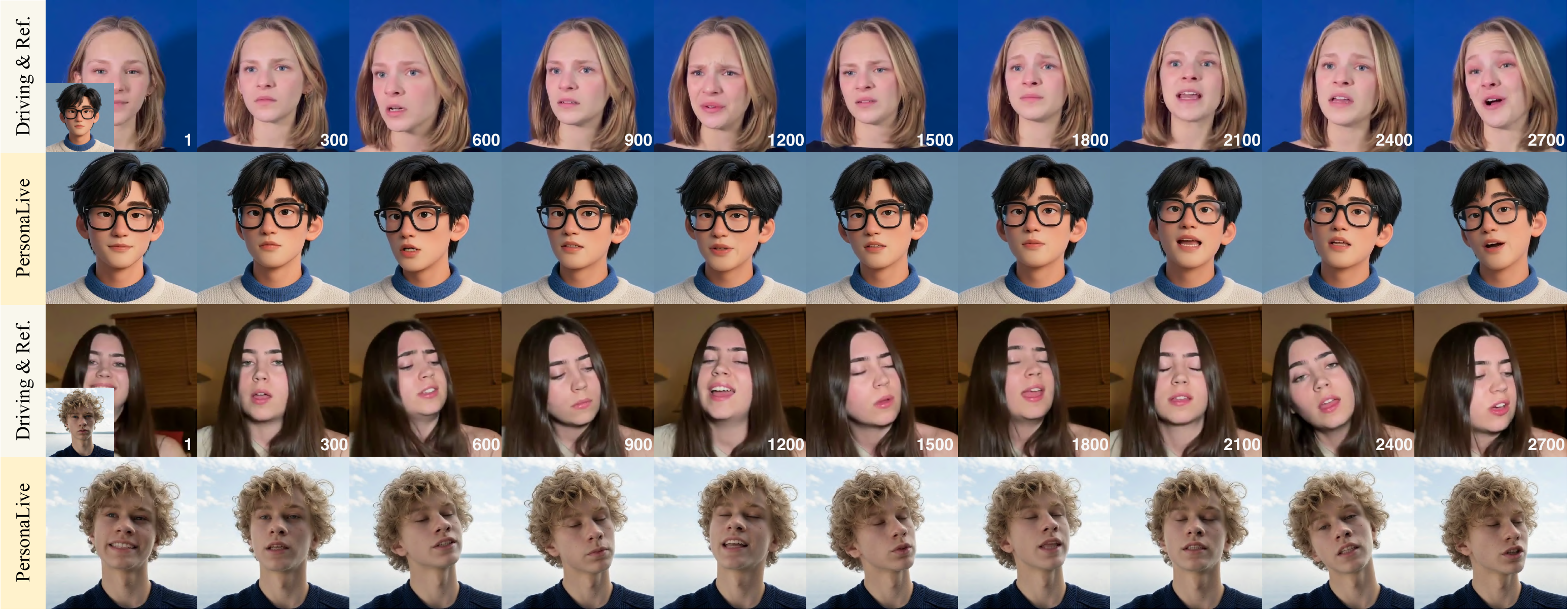}
    \vspace{-1em}
    \caption{Long avatar video results (3/4).
    }
    \label{fig:long_video_3}
\end{figure*}

\begin{figure*}[t]
    \centering
    \includegraphics[width=\textwidth]{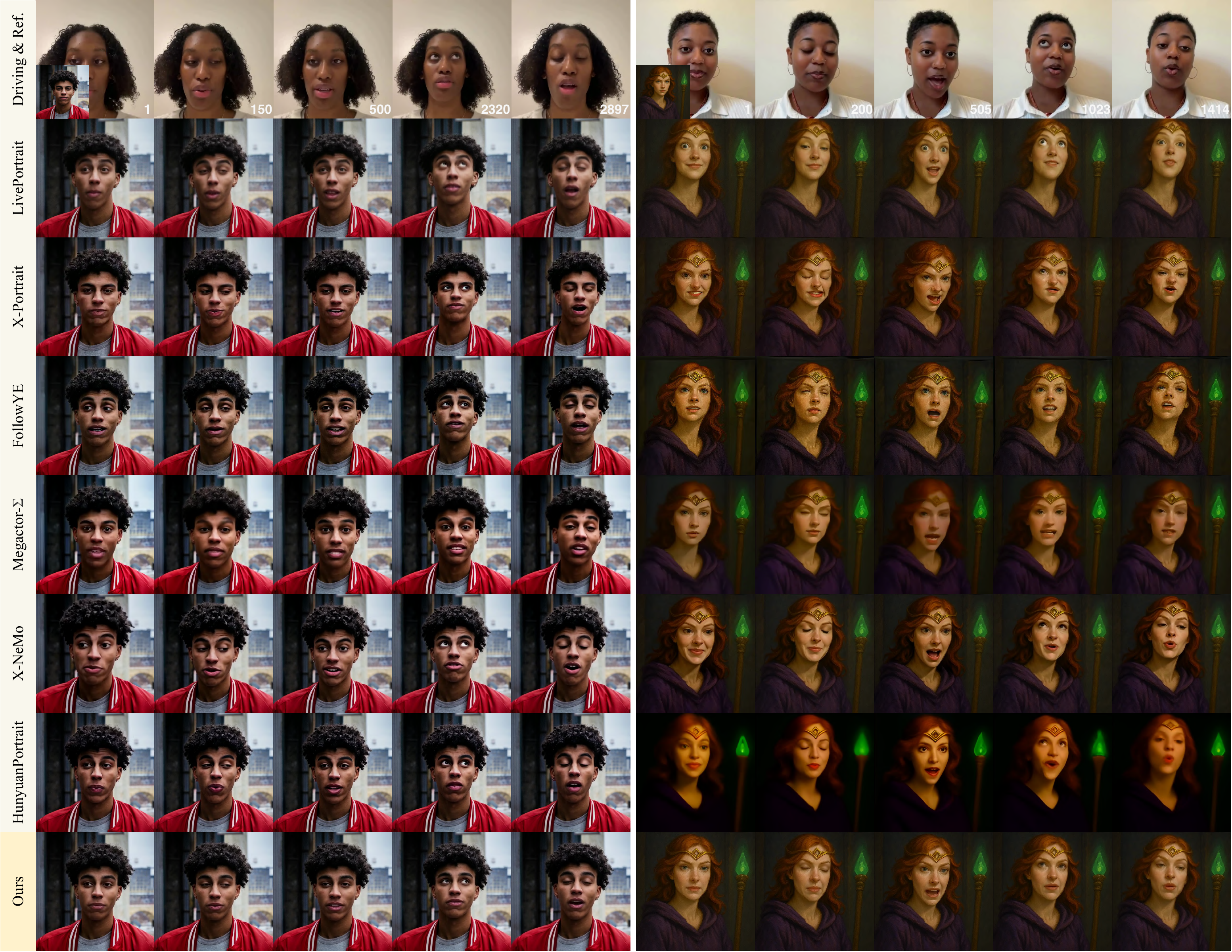}
    \vspace{-1em}
    \caption{More visualizations of cross-reenactment comparison (2/2).
    }
    \label{fig:more_cross_2}
\end{figure*}

\begin{figure*}[t]
    \centering
    \includegraphics[width=\textwidth]{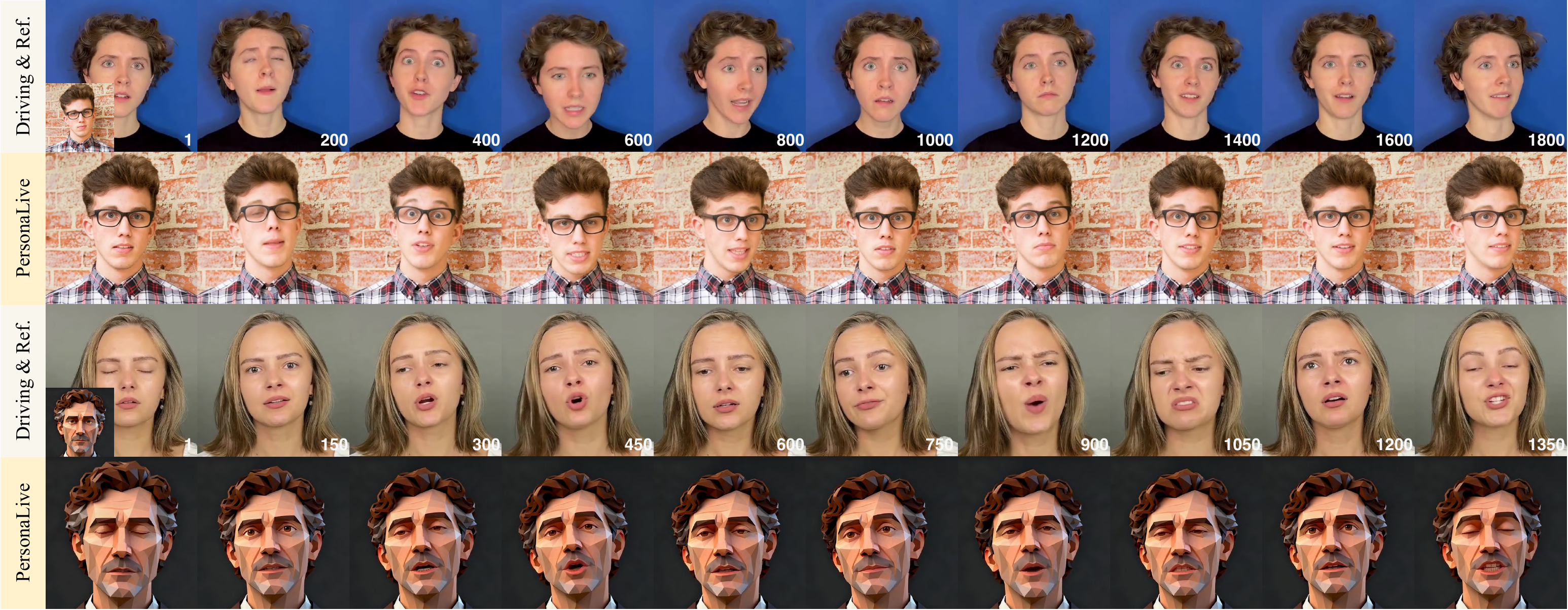}
    \vspace{-1em}
    \caption{Long avatar video results (4/4).
    }
    \label{fig:long_video_4}
\end{figure*}